\documentclass[letterpaper]{article} % DO NOT CHANGE THIS
\usepackage{aaai2027}
\usepackage[hyphens]{url}  % DO NOT CHANGE THIS
\usepackage{amsmath,amsfonts}
\usepackage{algorithmic}
\usepackage{algorithm}
\usepackage{array}
\usepackage[caption=false,font=normalsize,labelfont=sf,textfont=sf]{subfig}
\usepackage{textcomp}
\usepackage{url}
\usepackage{verbatim}
\usepackage{graphicx}
\usepackage{cite}
\usepackage{multirow}
\usepackage{xcolor}

\usepackage{longtable}
\usepackage{booktabs}
\usepackage{array}
\usepackage{tabularx} 
\usepackage{utfsym}
\usepackage{colortbl}
\usepackage{xcolor}
\usepackage{array}
\usepackage{color}

\usepackage{pifont}
\usepackage{natbib}  % DO NOT CHANGE THIS AND DO NOT ADD ANY OPTIONS TO IT
\usepackage{caption} % DO NOT CHANGE THIS AND DO NOT ADD ANY OPTIONS TO IT
\usepackage{newfloat}
\usepackage{listings}
\DeclareCaptionStyle{ruled}{labelfont=normalfont,labelsep=colon,strut=off} % DO NOT CHANGE THIS
\floatstyle{ruled}
\newfloat{listing}{tb}{lst}{}
\floatname{listing}{Listing}

\usepackage{booktabs}

\title{LabRobFail: A Benchmark for Robotic Failure Analysis in Chemical Self-driving Laboratory}

\author{
    Haobo Wang\textsuperscript{\rm 1,\rm 2}\equalcontrib,
    Baoli Sun\textsuperscript{\rm 1}\equalcontrib,
    Anqi Zou\textsuperscript{\rm 1,\rm 2},
    Dongsheng Huang\textsuperscript{\rm 1},
    Zelin Lv\textsuperscript{\rm 1},
    Ning Wang\textsuperscript{\rm 1},\\
    Rui Li\textsuperscript{\rm 3,\rm 4},
    Dongzhan Zhou\textsuperscript{\rm 4}\corresponding,
    Weiyu Guo\textsuperscript{\rm 5}\corresponding,
    Zhihui Wang\textsuperscript{\rm 1,\rm 2}\corresponding,
    Wanli Ouyang\textsuperscript{\rm 2,\rm 4,\rm 6}\corresponding
}
\affiliations{
    \textsuperscript{\rm 1}School of Software, Dalian University of Technology, Dalian, China\\
    \textsuperscript{\rm 2}Shenzhen Loop Area Institute-Hong Kong Science and Technology Innovation Cooperation Zone, Shenzhen, China\\
    \textsuperscript{\rm 3}School of Computer Science and Engineering, Harbin Institute of Technology, Harbin, China\\
    \textsuperscript{\rm 4}Shanghai AI Laboratory, Shanghai, China\\
    \textsuperscript{\rm 5}The Hong Kong University of Science and Technology, Hong Kong, China\\
    \textsuperscript{\rm 6}The Chinese University of Hong Kong, Hong Kong, China\\
    dongzhan.zhou@gmail.com, guoweiyu96@gmail.com, zhwang@dlut.edu.cn, wlouyang@ie.cuhk.edu.hk
}

\lstdefinelanguage{json}{
    morestring=[b]",
    morestring=[d]',
    literate=
        *{:}{{{\color{black}{:}}}}{1}
        {,}{{{\color{black}{,}}}}{1}
        {\{}{{{\color{black}{\{}}}}{1}
        {\}}{{{\color{black}{\}}}}}{1}
        {[}{{{\color{black}{[}}}}{1}
        {]}{{{\color{black}{]}}}}{1},
}

\begin{document}

\twocolumn[{
\renewcommand\twocolumn[1][]{#1}%
\maketitle
\vspace{-8pt}
\begin{center}
\includegraphics[width=1.0\textwidth]{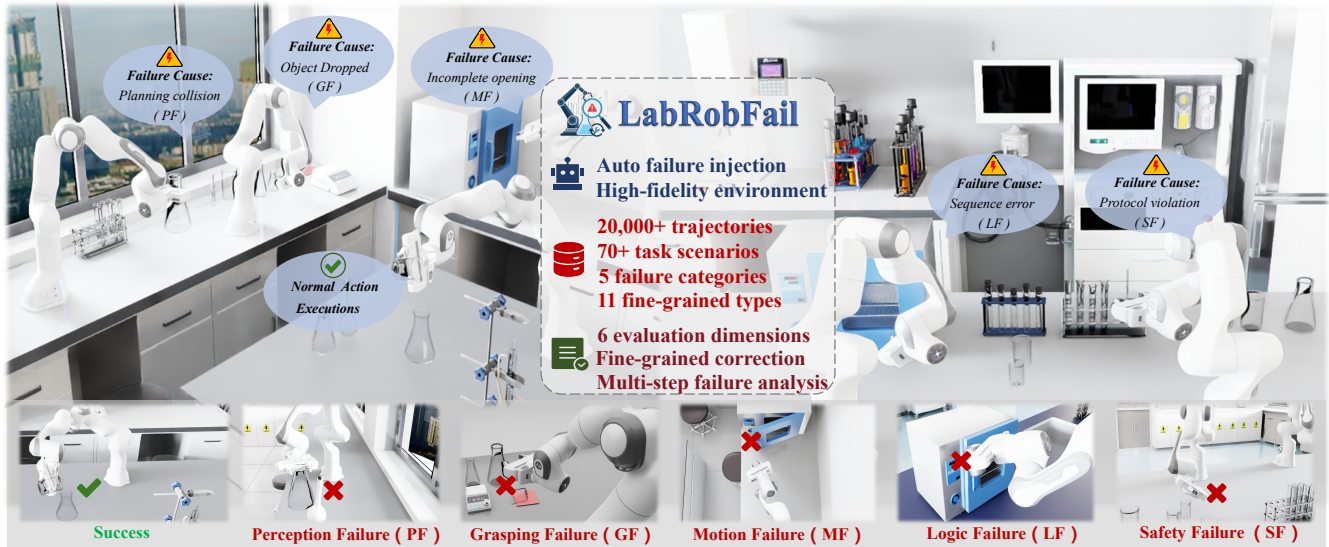}
\vspace{-10pt}
\captionof{figure}{\textbf{LabRobFail Framework for Chemical Laboratory Failure Analysis.} A high-fidelity simulation platform automatically injects failures across five categories: Perception (PF), Grasping (GF), Motion (MF), Logic (LF), and Safety (SF). The resulting data support a six-dimensional benchmark for failure detection, localization, diagnosis, and fine-grained correction.}
\label{fig:data1}
\end{center}%
\vspace{-10pt}
}]

% \maketitle
% \begin{figure*}[!h]
% \centering
% \vspace{-4pt}
% % \fbox{\rule{0pt}{3.6in} \rule{0.99\linewidth}{0pt}}
% \centerline{\includegraphics[width=1\linewidth]{figs/motiva.pdf}}
% \caption{\textbf{LabRobFail Framework for Chemical Laboratory Failure Analysis.} A high-fidelity simulation platform automatically injects failures across five categories: Perception (PF), Grasping (GF), Motion (MF), Logic (LF), and Safety (SF). The resulting data support a six-dimensional benchmark for failure detection, localization, diagnosis, and fine-grained correction.}
% \label{fig:data1}
% \vspace{-4pt}
% \end{figure*}

\begin{abstract}
The deployment of embodied agents in self-driving laboratories could accelerate scientific discovery, yet their reliability is constrained by the irreversible and safety-critical nature of chemical experiments. Progress is further hindered by scarce failure data and the lack of fine-grained evaluation protocols. To address these challenges, we introduce LabRobFail, a failure-centric framework for learning and evaluating robotic failure analysis in chemical laboratories. LabRobFail-Sim injects controllable failures at the control, physics, and semantic levels, enabling the construction of LabRobFail-Data, which contains over 20,000 trajectories across 70+ task scenarios, five failure categories, and 11 fine-grained failure types. LabRobFail-Bench evaluates six capabilities spanning task understanding, failure detection, temporal localization, severity assessment, failure classification, and actionable correction. We further develop LabRobFail-VLM, a domain-specialized vision-language model that generates structured failure diagnoses and recovery instructions. On seen environments, it achieves 90.83\% failure-detection accuracy and 77.21\% temporal-localization accuracy, substantially outperforming general-purpose VLMs. When integrated as a real-time supervisor, it improves downstream task success rates by 4–16 percentage points, demonstrating the value of fine-grained failure understanding for closed-loop recovery and reliable laboratory autonomy.
 
\end{abstract}
\section{Introduction}

\begin{table*}[t]
\centering
\caption{Comparison of robotic failure detection datasets and benchmarks. Lab Env. / Lab Inst.: Support for chemical laboratory environments and specialized laboratory instruments. Auto Inj.: Automated failure injection. \# Fail. Types / \# Eval Dims: Number of failure categories and evaluation dimensions. Fine Corr. / Temp. Loc. / Severity: Fine-grained correction, temporal localization, and severity assessment capabilities. The \textbf{best} and \underline{second-best} results are highlighted.}
\label{tab:dataset_comparison}
\resizebox{\textwidth}{!}{
\begin{tabular}{l|cc|cccc|cccc}
\toprule
\multirow{2}{*}{\textbf{Dataset/Benchmark}} & \multicolumn{2}{c|}{\textbf{Laboratory Support}} & \multicolumn{4}{c|}{\textbf{Data Statistics}} & \multicolumn{4}{c}{\textbf{Detection \& Correction Capability}} \\
\cmidrule(lr){2-3} \cmidrule(lr){4-7} \cmidrule(lr){8-11}
& Lab Env. & Lab Inst. & Auto Inj. & \# Failure Types & \# Eval Dims & \# Traj. & Fine Corr. & Temp. Loc. & Severity \\
\midrule
ARMBench Video Defect~\cite{mitash2023armbench} & \ding{55} & \ding{55} & \ding{55} & 2 & 1 & 4,070 & \ding{55} & \ding{55} & \ding{55} \\
ViFailback dataset~\cite{zeng2025diagnose} & \ding{55} & \ding{55} & \ding{55} & 4 & \underline{2} & 5,202 & \ding{55} & \ding{55} & \ding{55} \\
RLBench-Fail~\cite{pacaud2025guardian} & \ding{55} & \ding{55} & \checkmark & 5 & 1 & 14,358 & \ding{55} & \ding{55} & \ding{55} \\
BridgeDataV2-Fail~\cite{pacaud2025guardian} & \ding{55} & \ding{55} & \checkmark & 5 & 1 & 9,830 & \ding{55} & \ding{55} & \ding{55} \\
UR5-Fail~\cite{pacaud2025guardian} & \ding{55} & \ding{55} & \checkmark & 5 & 1 & 570 & \ding{55} & \ding{55} & \ding{55} \\
SMF-DROID~\cite{grislain2025failsense} & \ding{55} & \ding{55} & \checkmark & 1 & \underline{2} & 6,276 & \ding{55} & \ding{55} & \ding{55} \\
RoboFAC~\cite{lu2025robofac} & \ding{55} & \ding{55} & \checkmark & 6 & \underline{2} & 10,722 & \ding{55} & \ding{55} & \ding{55} \\
FAILURE~\cite{thoduka2024multimodal} & \ding{55} & \ding{55} & \ding{55} & 5 & \underline{2} & 229 & \ding{55} & \ding{55} & \ding{55} \\
AHA~\cite{duan2024aha} & \ding{55} & \ding{55} & \checkmark & \underline{7} & \underline{2} & \textbf{49K} & \ding{55} & \ding{55} & \ding{55} \\
\midrule
\textbf{LabRobFail-Data (Ours)} & \checkmark & \checkmark & \checkmark & \textbf{11} & \textbf{6} & \underline{20K} & \checkmark & \checkmark & \checkmark \\
\bottomrule
\end{tabular}
}
\vspace{-8pt}
\end{table*}

% The realization of Self-Driving Laboratories (SDLs)~\cite{szymanski2023autonomous,abolhasani2023rise,lan2025autobio,li2025labutopia} represents a pivotal leap in accelerating scientific discovery, promising to automate tedious experimentation through intelligent embodied agents. Recent advances in Vision-Language-Action (VLA) models~\cite{zitkovich2023rt,kim2024openvla,zhen20243d,black2025pi05} have demonstrated impressive capabilities in generalist manipulation tasks. However, a critical reliability bottleneck emerges when deploying these agents into the high-stakes environment of a chemical laboratory. Unlike household settings~\cite{liu2023reflect,duan2024aha,pacaud2025guardian} where errors are often reversible (e.g., dropping a fruit), scientific experiments are governed by strict protocols with irreversible outcomes and safety hazards. Even slight execution deviation, such as misaligning a pipette or agitating a volatile reagent too aggressively, can result in the contamination of rare samples, invalidation of long-horizon workflows, or catastrophic safety breaches. Consequently, the transition from automation to true autonomy demands not merely the capability to execute tasks, but the intelligence~\cite{qu2024recursive,lin2023learning,kambhampati2024position} to perceive, diagnose, and recover from failures in real-time.
The emergence of Self-Driving Laboratories (SDLs)~\cite{szymanski2023autonomous,lan2025autobio,li2025labutopia} offers a promising route to accelerate scientific discovery through embodied automation. Although recent Vision-Language-Action (VLA) models~\cite{kim2024openvla,zhen20243d,black2025pi05} have shown strong generalist manipulation capabilities, their deployment in chemical laboratories remains constrained by reliability. Unlike household tasks, where failures are often reversible~\cite{liu2023reflect,duan2024aha,pacaud2025guardian}, laboratory experiments follow strict protocols and involve irreversible processes and substantial safety risks. Minor deviations, such as pipette misalignment or excessive agitation, may contaminate samples, invalidate long-horizon workflows, or cause hazardous incidents. Therefore, reliable laboratory autonomy requires agents not only to execute tasks, but also to perceive, diagnose, and recover from failures in real time~\cite{qu2024recursive,lin2023learning,kambhampati2024position}.

While recent studies have applied Vision-Language Models (VLMs) to robotic failure detection~\cite{guo2024doremi,duan2024aha,zhou2025code}, their extension to self-driving laboratories is limited by two key gaps.
\textbf{(1) The scarcity of large-scale failure data.} Real-world failure collection is costly and hazardous, while existing scientific simulators such as LabUtopia~\cite{li2025labutopia} and AutoBio~\cite{lan2025autobio} primarily emphasize successful execution and lack systematic failure-injection pipelines. Consequently, current agents are trained largely on successful trajectories, leaving insufficient coverage of diverse laboratory anomalies.
\textbf{(2) The coarseness of existing evaluation benchmarks.} Existing benchmarks, largely designed for household manipulation~\cite{duan2024aha,thoduka2024multimodal}, mainly assess binary detection or coarse failure categories, overlooking fine-grained localization, classification, and severity assessment. Although some methods provide corrective feedback~\cite{duan2024aha,lu2025robofac}, it is often too ambiguous for direct execution. This is particularly problematic in irreversible chemical procedures, which require precise, risk-aware recovery rather than trial-and-error.

To address these gaps, we introduce LabRobFail, a failure-centric framework for robotic failure analysis in self-driving laboratories, comprising a simulation platform, a large-scale dataset, and a multi-dimensional benchmark.
First, building upon the physical realism established by LabUtopia, we develop \textbf{LabRobFail-Sim}, which extends these environments with a novel automated failure injection pipeline. 
Instead of relying on manually designed faults, LabRobFail-Sim systematically perturbs control signals, physical dynamics, and task-level conditions to generate large-scale annotated failure trajectories covering execution, perception, and semantic anomalies.
Based on this pipeline, we construct \textbf{LabRobFail-Data}, comprising 5 major failure categories (Perception, Grasping, Motion, Logic, and Safety), 11 fine-grained failure types, and 70+ task instances ranging from short-horizon manipulations to long-horizon multi-stage workflows, totaling over 20,000 operation trajectories. 
Crucially, the dataset balances successful executions with diverse failure modes, enabling models to learn the precise discriminative boundary between normal operations and subtle anomalies. Finally, we design \textbf{LabRobFail-Bench}, the first multi-dimensional failure understanding benchmark tailored for chemical laboratories. LabRobFail-Bench provides systematic evaluation covering failure detection,  localization, mode classification, severity assessment, and fine-grained correction planning. It directly addresses the stringent safety requirements of laboratory automation, paving the way for closed-loop recovery in high-stakes scientific experiments.

% Building upon LabRobFail-Data, we introduce LabRobFail-VLM, a vision-language model fine-tuned for laboratory failure analysis. LabRobFail-VLM detects failures, determines their types and severity, and generates correction instructions to guide recovery operations. Experiments show that LabRobFail-VLM outperforms general-purpose VLMs such as GPT-5.4 across all dimensions, and when integrated as an external supervisor, significantly improves robotic policy success rates by timely intervention before failures cascade into catastrophes. In summary, our contributions are three-fold: 

% Building upon LabRobFail-Data, we introduce LabRobFail-VLM, which casts laboratory failure analysis as a unified structured prediction problem over failure existence, temporal localization, failure type, severity, and corrective action guidance. This formulation moves beyond conventional failure detection as a single-label recognition task, and instead equips a vision-language model with the ability to produce semantically grounded and recovery-oriented failure descriptions for scientific robotic systems. Under this formulation, LabRobFail-VLM serves not merely as a domain-finetuned baseline, but as a task-specific supervisory model that connects failure understanding with actionable recovery. Empirically, it consistently surpasses general-purpose VLMs such as GPT-5.4 across benchmark dimensions, and, when integrated into the control loop, provides initial evidence of improved downstream policy recovery.
Building on LabRobFail-Data, we introduce LabRobFail-VLM, which formulates laboratory failure analysis as structured prediction over failure presence, temporal localization, type, severity, and corrective guidance. Beyond single-label detection, it produces semantically grounded, recovery-oriented diagnoses and serves as a task-specific supervisor linking failure understanding to actionable recovery. Experiments show that LabRobFail-VLM consistently outperforms general-purpose VLMs  and improves downstream policy recovery when integrated into the control loop.

In summary, 
(1) We introduce LabRobFail-Sim, a failure-centric simulation framework that enables scalable and controllable failure generation through automated perturbations at the control, physics, and semantic levels.
(2) We construct LabRobFail-Data and LabRobFail-Bench, establishing the first large-scale dataset and multi-dimensional evaluation standard for failure reasoning in chemical laboratory environments.
(3) We develop LabRobFail-VLM, demonstrating that domain-specialized failure analysis can support actionable diagnosis and closed-loop recovery.

\section{Related Work}
\subsection{Embodied AI and Simulation for SDLs}

\begin{figure*}[!h]
\centering
% \fbox{\rule{0pt}{3.6in} \rule{0.99\linewidth}{0pt}}
\centerline{\includegraphics[width=1\linewidth]{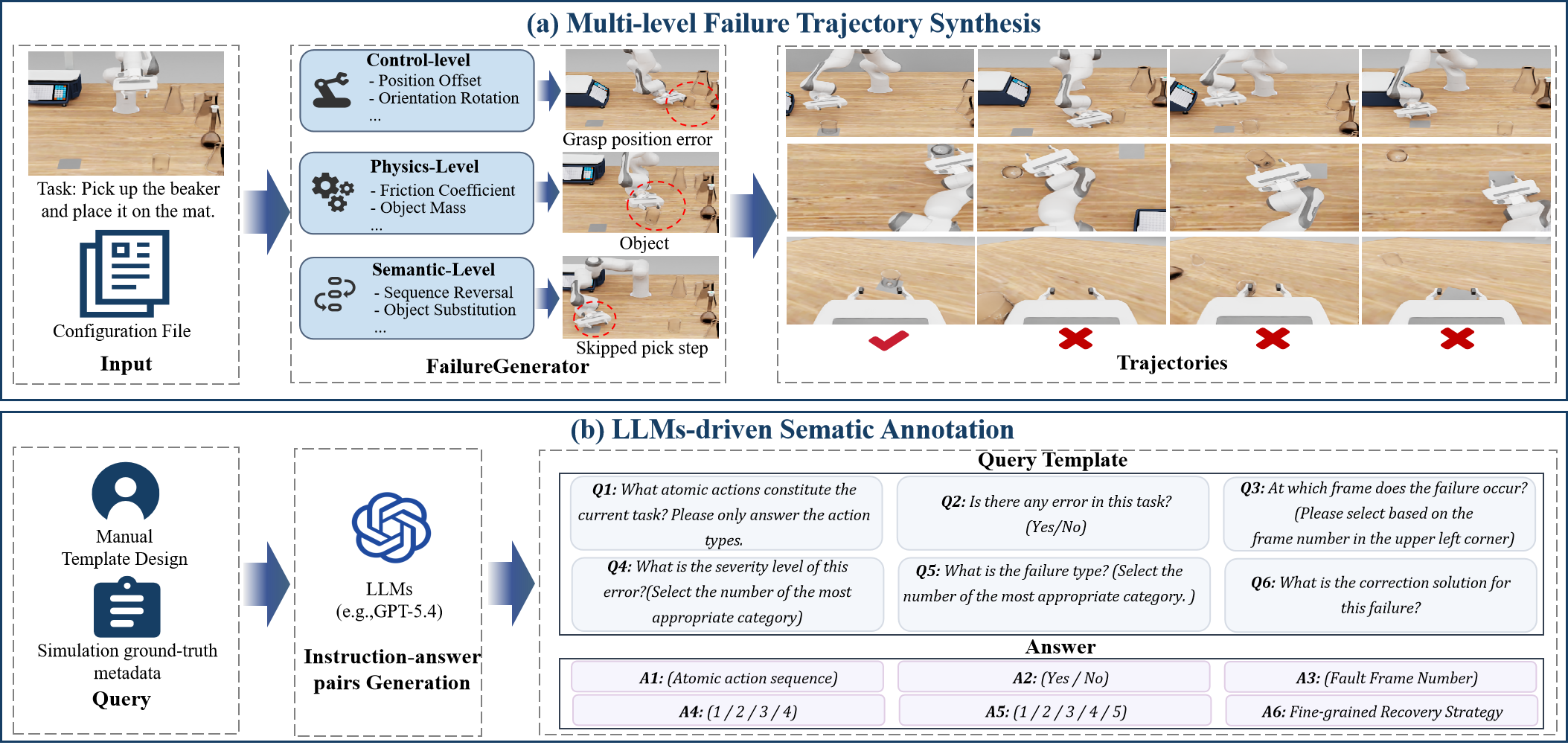}}
\vspace{-4pt}
\caption{\textbf{The LabRobFail-Sim Framework.} (a) The multi-level failure generator injects faults via Control, Physics, and Semantic perturbations. (b) The LLM-driven semantic annotation pipeline leverages simulation metadata and GPT-5.4 to automatically generate instruction-answer pairs.} 
\label{fig:inject}
\vspace{-10pt}
\end{figure*} 

% The paradigm of scientific automation is shifting from rigid, script-based systems toward intelligent embodied agents. Early works like Chemputer~\cite{steiner2019organic} and Artificial Chemist~\cite{epps2020artificial} laid the groundwork by standardizing chemical synthesis through programming languages and closed-loop optimization. However, developing generalist agents requires diverse training environments beyond physical hardware. To this end, high-fidelity simulators have emerged as critical infrastructures. AutoBio~\cite{lan2025autobio} introduced specialized physics plugins for biological protocols, focusing on the precision of micro-manipulation. Concurrently, LabUtopia~\cite{li2025labutopia} established a comprehensive benchmark for chemical experiments, supporting multi-physics interactions across 30 distinct tasks. Despite these advancements, a critical gap remains: existing platforms are primarily designed to demonstrate task feasibility (success). They lack systematic mechanisms for failure injection, resulting in a "survivorship bias" where agents are trained only on ideal trajectories. LabRobFail addresses this by introducing a controllable anomaly generation pipeline specifically designed to model the irreversible and hazardous failures unique to scientific workflows.
Scientific automation is evolving from rigid, script-based systems toward  embodied agents. While recent works like Chemputer~\cite{steiner2019organic} and Artificial Chemist~\cite{epps2020artificial} standardized synthesis via closed-loop optimization, training generalist agents requires high-fidelity simulators beyond physical hardware. Recent platforms such as AutoBio~\cite{lan2025autobio} and LabUtopia~\cite{li2025labutopia} have advanced this frontier by simulating biological micro-manipulation and multi-physics chemical interactions, respectively. However, these environments focus on successful execution and lack systematic failure injection, leading to a survival bias toward ideal trajectories. LabRobFail addresses this gap by introducing a controllable anomaly generation pipeline for safety-critical failures in scientific workflows.

% \subsection{Robotic Failure Detection Datasets}

% The growing adoption of vision-language models for robotic failure analysis has driven the construction of specialized datasets. Early efforts focused on binary failure detection: REFLECT~\cite{liu2023reflect} introduced the RoboFail dataset with LLM-generated failure explanations, enabling robots to summarize past experiences and query language models for failure reasoning. AHA~\cite{duan2024aha} developed the FailGen framework for procedural failure trajectory generation by perturbing successful demonstrations, establishing the first large-scale robotic failure dataset with 49K failure-question pairs. Subsequent works expanded toward finer-grained analysis. RoboFAC~\cite{lu2025robofac} proposed a hierarchical three-level failure taxonomy spanning task planning, motion planning, and execution control, collecting 9,440 erroneous trajectories across both simulated and real-world environments. Guardian~\cite{guo2025guardian} constructed RLBench-Fail and BridgeDataV2-Fail datasets with fine-grained failure categorization across both simulation and real-world environments. 

% However, these benchmarks target household scenarios, neglecting the unique demands of chemical laboratories that require precise failure understanding and fine-grained correction. LabRobFail-Bench addresses this gap through six evaluation dimensions covering failure detection, localization, classification, severity assessment, and correction planning, providing the first systematic failure analysis framework for self-driving laboratories.

\subsection{Robotic Failure Understanding and Benchmarks}
The integration of VLMs has shifted robotic failure analysis from fixed sensor thresholds toward semantic reasoning. REFLECT~\cite{liu2023reflect} used LLMs for retrospective log-based diagnosis, while AHA~\cite{duan2024aha} introduced FailGen to generate large-scale failure data through procedural perturbations. Later studies improved failure granularity: RoboFAC~\cite{lu2025robofac} developed a hierarchical taxonomy of planning and execution errors, and Guardian~\cite{pacaud2025guardian} extended failure categorization across multiple manipulation benchmarks~\cite{james2020rlbench,walke2023bridgedata}. However, existing benchmarks remain centered on household settings with largely reversible failures and provide limited support for precise recovery. LabRobFail-Bench addresses this gap by evaluating failure detection, localization, severity assessment, and fine-grained correction in safety-critical chemical laboratories.

\section{LabRobFail}

% We introduce LabRobFail, a unified framework for robotic failure analysis in chemical laboratories. It comprises \textbf{LabRobFail-Sim}, a simulator with automated failure injection; \textbf{LabRobFail-Data}, containing 20,000+ trajectories across 5 categories; and \textbf{LabRobFail-Bench}, a multi-dimensional benchmark for failure analysis and recovery.

% \begin{figure*}[!h]
% \centering
% % \fbox{\rule{0pt}{3.6in} \rule{0.99\linewidth}{0pt}}
% \centerline{\includegraphics[width=1\linewidth]{figs/inject.png}}
% \caption{LabRobFail-Data construction and LabRobFail-VLM application pipeline. (a) LabRobFail-Sim, Automated Failure Data Generation: three-level failure injection (Control, Physics, Semantic) with GPT-5.4-assisted annotation for LabRobFail-Data construction. (b) LabRobFail-VLM Training: Qwen3-VL is fine-tuned for task understanding, failure detection, and correction reasoning. (c) VLA Correction & Recovery: fine-grained correction instructions enable VLA models to recover from failures.} 
% \label{fig:inject}
% % \vspace{-10pt}
% \end{figure*} 

\subsection{LabRobFail-Sim}
% Building upon NVIDIA Isaac Sim, LabRobFail-Sim inherits laboratory scene assets and physical configurations from existing simulation frameworks. As illustrated in Fig.~\ref{fig:inject} (a), we introduce an automated failure data generation pipeline comprising two modules: multi-level failure trajectory synthesis and GPT-5.4-assisted annotation, enabling scalable LabRobFail-Data construction.
% While recent environments like LabUtopia~\cite{li2025labutopia} provide high-fidelity physical interactions for chemical experiments, they are primarily designed to validate successful task execution. To bridge the gap between ideal simulations and the hazardous stochasticity of real-world laboratories, we develop LabRobFail-Sim.
% Constructed upon the physics engine of LabUtopia, LabRobFail-Sim introduces a novel automated failure injection pipeline. This pipeline comprises two synergistic modules: multi-level failure trajectory synthesis and LLM-driven semantic annotation, collectively enabling the scalable construction of LabRobFail-Data.
Although LabUtopia~\cite{li2025labutopia} provides high-fidelity simulation of chemical interactions, it primarily focuses on successful task execution. Built on its physics engine, we develop LabRobFail-Sim to model the stochastic failures of real laboratories through an automated pipeline comprising multi-level failure trajectory synthesis and LLM-driven semantic annotation. Together, these modules enable the scalable generation and annotation of LabRobFail-Data.

% \subsubsection{Multi-level Failure Trajectory Synthesis.} 
% \textbf{Multi-level Failure Trajectory Synthesis.} To facilitate controllable generation, we formulate robot manipulation skills as sequences of sparse keyframes, each defining the end-effector's target state (rotation and translation) and gripper status. Based on this representation, we design a FailureGenerator module that performs systematic perturbations at three levels: (1) Control-level perturbation modifies keyframe parameters including position offsets, orientation rotations, and gripper command corruption. (2) Physics-level perturbation dynamically adjusts simulation parameters such as friction coefficients and object mass to induce emergent failures like slippage or grasp instability. (3) Semantic-level perturbation introduces task logic errors including operation sequence reversal, target object substitution, and safety constraint violations. To enable scalable data generation, our pipeline adopts a configuration-driven design where each task is associated with configuration files specifying perturbable keyframes and parameter ranges. The perturbed keyframes are then injected into the simulator to generate failure trajectories with automatic outcome labeling.

\subsubsection{Multi-level Failure Trajectory Synthesis} 
In Fig.~\ref{fig:inject} (a), to facilitate controllable generation, we formulate a robot manipulation skill as a sequence of sparse keyframes $\tau = \{ (T_i, g_i, o_i)\}_{i=1}^{N}$, where $T_i = (R_i, p_i) \in SE(3)$ denotes the end-effector's target pose (rotation matrix $R_i$ and translation vector $p_i$), $g_i \in \{0,1\}$ represents the gripper status, and $o_i$ is the interaction target. 
The \textbf{FailureGenerator} module synthesizes failure trajectories $\tilde{\tau}$ by applying systematic perturbation functions $\Phi$ across three hierarchical levels. 

\textbf{(1) Control-level Perturbation ($\Phi_{ctrl}$):} We simulate execution errors by injecting stochastic noise into keyframe parameters. Specifically, we apply Gaussian perturbation to the translation $p_i$ and Lie algebra noise to the rotation $R_i$:
    \begin{equation}
        \label{eq:control_perturb}
        \tilde{p}_i = p_i + \xi_{trans}, \quad \tilde{R}_i = R_i \cdot \text{Exp}(\xi_{rot}),
    \end{equation}
    where $\xi \sim \mathcal{N}(0, \Sigma)$. Gripper commands are corrupted with a failure probability $\lambda_{grip}$ to model actuator faults.

\begin{figure*}[!h]
\centering
% \fbox{\rule{0pt}{3.6in} \rule{0.99\linewidth}{0pt}}
\centerline{\includegraphics[width=1\linewidth]{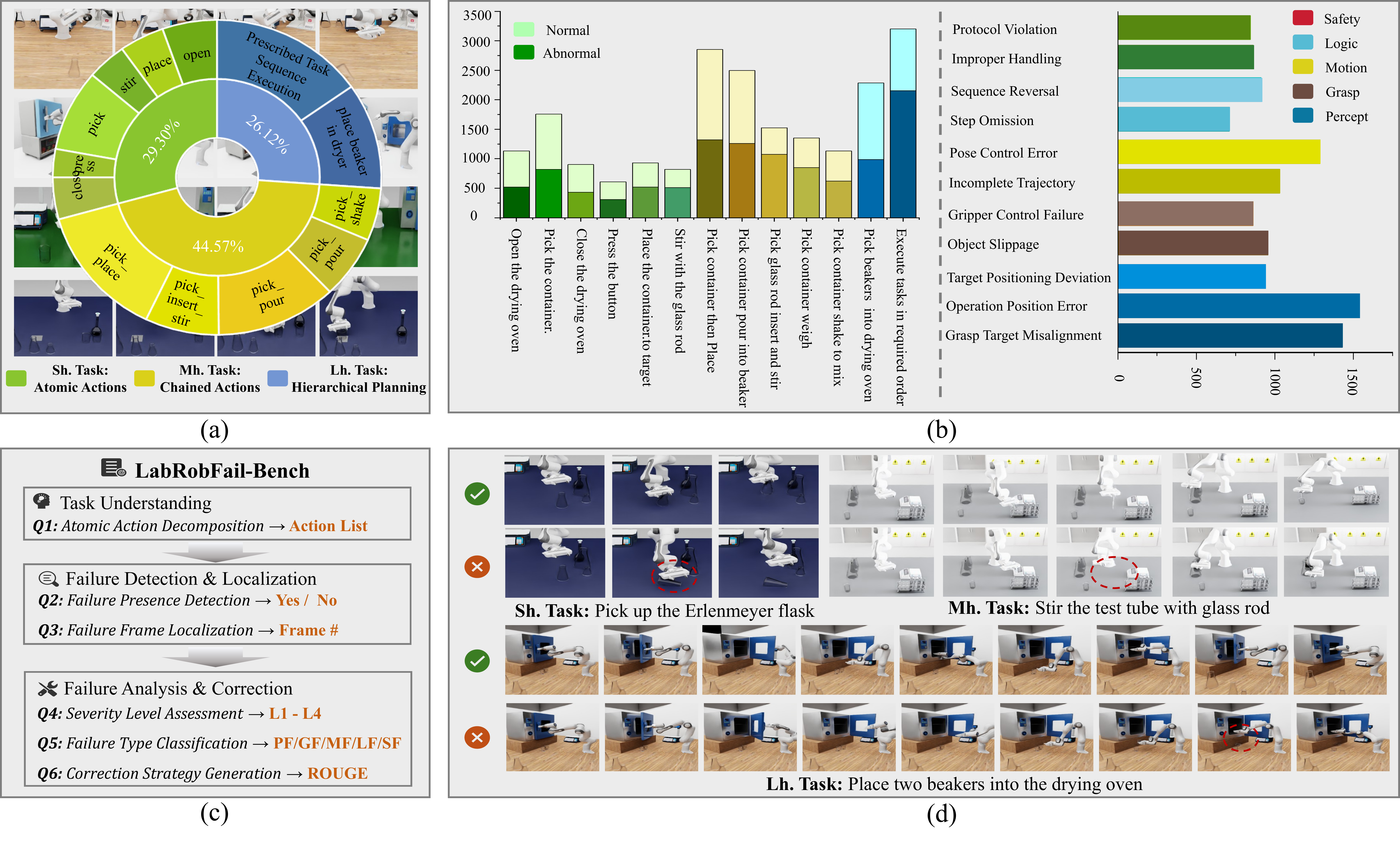}}
\vspace{-4pt}
\caption{\textbf{LabRobFail-Data and LabRobFail-Bench.} (a) Distribution of short-, medium-, and long-horizon tasks, corresponding to atomic actions, chained actions, and hierarchical planning. (b) Distribution of normal and failed trajectories across tasks (left), and 11 fine-grained failure types grouped into five categories (right). (c) Six-dimensional benchmark covering task understanding, failure detection and localization, and failure analysis and correction. (d) Representative success–failure trajectory pairs at each task complexity level.}
\label{fig:data}
\vspace{-10pt}
\end{figure*}

\textbf{(2) Physics-level Perturbation ($\Phi_{phy}$):} To induce emergent failures like slippage or grasp instability, we dynamically adjust the simulation dynamics parameters $\Psi = \{ \mu, m, \nu, \dots \}$ (representing friction, mass, viscosity, etc.). The perturbed parameters $\tilde{\Psi}$ are derived via uniform scaling:
    \begin{equation}
        \label{eq:physics_perturb}
        \tilde{\psi} = \psi \cdot (1 + \delta), \quad \text{with } \delta \sim \mathcal{U}(-\alpha, \alpha).
    \end{equation}
    This ensures that the environment dynamics deviate from the nominal model, testing the agent's physical robustness.

\textbf{(3) Semantic-level Perturbation ($\Phi_{sem}$):} This level introduces task logic errors. We model this as a permutation operator $\pi$ on the sequence indices to simulate step reversals or skips ($\tilde{\tau}_{seq} = \{ T_{\pi(1)}, \dots \}$), and a mapping function $\mathcal{M}(o_i)$ for target object substitution (e.g., selecting the wrong reagent), leading to safety constraint violations.

% Formally, the final failure trajectory is synthesized as $\tilde{\tau} = \Phi_{sem}( \Phi_{ctrl}(\tau) )$ executed under the perturbed dynamics $\Phi_{phy}(\Psi)$. To enable scalable data generation, our pipeline adopts a \textbf{configuration-driven design} where each task is associated with configuration files specifying perturbable keyframes and parameter ranges $\alpha$. The perturbed keyframes are then injected into the simulator to generate failure trajectories with automatic outcome labeling.
The trajectory $\tilde{\tau} = \Phi_{sem}( \Phi_{ctrl}(\tau) )$ is executed under perturbed dynamics $\Phi_{phy}(\Psi)$. Adopting a scalable \textbf{configuration-driven design}, we specify perturbable keyframes and ranges $\alpha$ to generate labeled trajectories.

\subsubsection{LLM-driven semantic annotation} 
Because manual annotation of temporal and semantic labels is impractical at scale, we develop a GPT-5.4-based VQA annotation pipeline, as shown in Fig.~\ref{fig:inject}(b). Using simulation ground-truth metadata, it instantiates query and correction templates aligned with the six dimensions of LabRobFail-Bench to generate context-specific instruction--answer pairs. The annotations are then validated through rule-based filtering and human sampling before being aligned with failure trajectories to construct LabRobFail-Data. Prompting and verification details are provided in the Supplementary Material.

\subsection{LabRobFail-Data}

Based on LabRobFail-Sim, we construct LabRobFail-Data, a large-scale dataset for laboratory robotic failure analysis in chemical laboratories. It comprises over \textbf{20,000} interaction trajectories across \textbf{70+} distinct laboratory task scenarios.

% As shown in Fig.~\ref{fig:data}(a), tasks are organized into three complexity levels: short-horizon tasks (Sh. Task, $\leq$10 frames) focus on atomic operations such as pick, place, and pour; medium-horizon tasks (Mh. Task, 10-30 frames) chain 2-3 atomic operations to examine failure detection during action transitions; and long-horizon tasks (Lh. Task, $\geq$30 frames) involve multi-stage workflows requiring hierarchical planning and precise temporal fault localization. To facilitate learning on complex scenarios, we oversample medium and long tasks to ensure sufficient coverage for multi-step generalization.
\textbf{Task Complexity Levels.}
As shown in Fig.~\ref{fig:data}(a), we organize experiment tasks into three complexity levels: \textbf{(1) Short-horizon Tasks (Sh. Task, $\leq$10 frames):} Focus on atomic operations such as \textit{pick}, \textit{place}, and \textit{pour}. Failures here are typically instantaneous execution errors. \textbf{(2) Medium-horizon Tasks (Mh. Task, 10-30 frames):} Chain 2-3 atomic operations (e.g., \textit{stir the beaker with glass rod}) to examine failure detection during critical action transitions. \textbf{(3) Long-horizon Tasks (Lh. Task, $>$50 frames):} Involve complex multi-stage workflows (e.g., \textit{pick beaker in dryer}) that require hierarchical planning and precise fault localization. We oversample the latter two levels to ensure sufficient coverage of multi-step workflows.

% The failure taxonomy is illustrated in Fig.~\ref{fig:data}(b). We design a systematic taxonomy covering 5 major categories and 11 fine-grained failure types, spanning perception, control, execution, planning, and safety aspects of laboratory robotics. Perception Failure (PF) arises from spatial localization errors, including target positioning deviation and incorrect placement locations. Grasping Failure (GF) reflects end-effector control deficiencies, such as object slippage due to insufficient force or gripper malfunction. Motion Failure (MF) corresponds to trajectory execution anomalies, including incomplete paths and pose control errors. Logic Failure (LF) captures task planning errors, such as step omission or sequence reversal. Safety Failure (SF) includes improper reagent handling and protocol violations. While such failures may have limited impact in general scenarios, they can lead to hazardous material leakage, equipment damage, or even personnel injury in chemical laboratory environments.

\begin{figure*}[t]
\centering
\includegraphics[width=\linewidth]{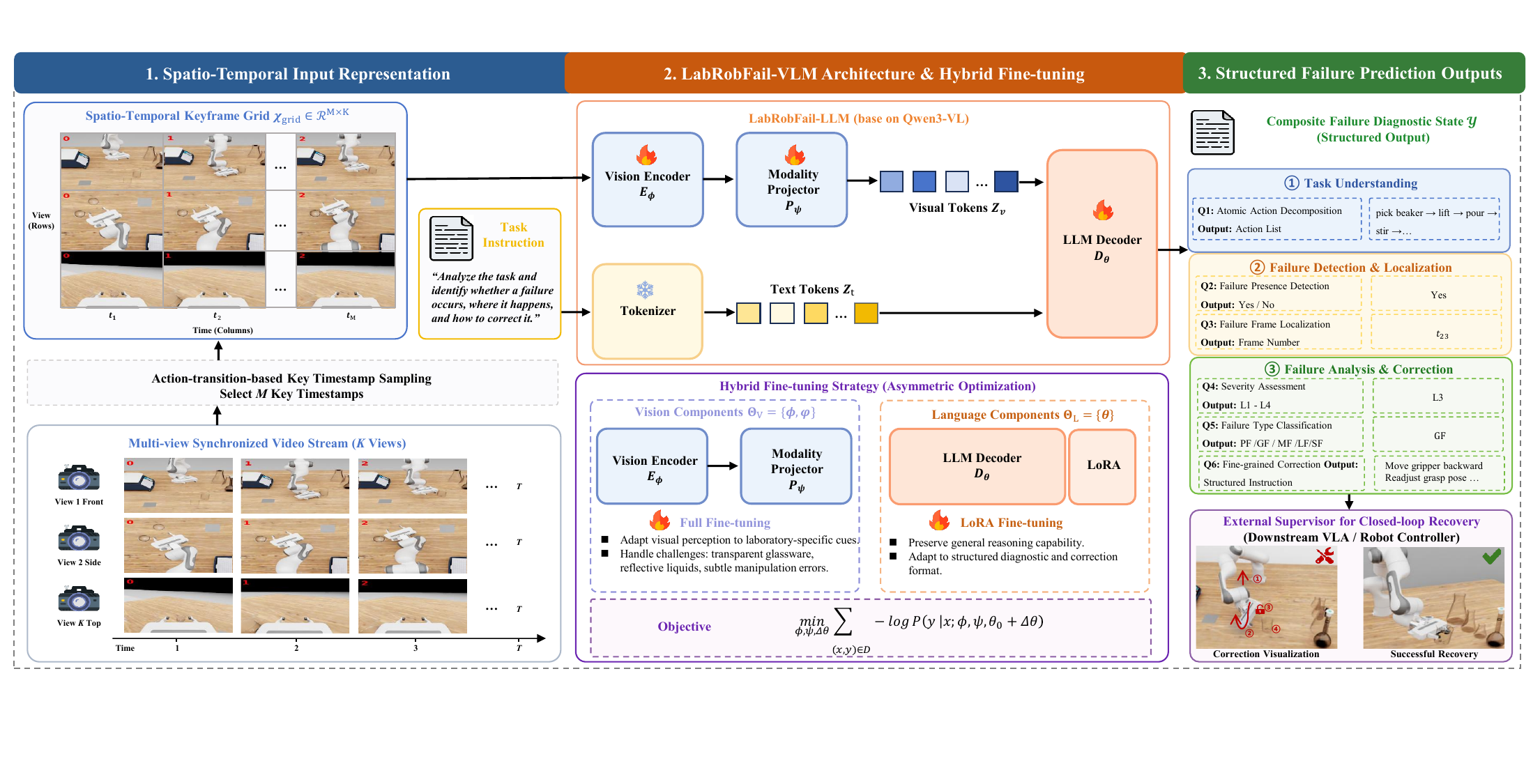}
\vspace{-45pt}
\caption{\textbf{Overview of LabRobFail-VLM.} Temporally indexed multi-view keyframes are processed by a hybrid-tuned Qwen3-VL to jointly perform failure detection, localization, diagnosis, and correction.}
\label{fig:vlm}
\vspace{-10pt}
\end{figure*}

\textbf{Fine-grained Failure Taxonomy.}
We design a systematic taxonomy covering \textbf{5 major categories} and \textbf{11 fine-grained types} (Fig.~\ref{fig:data}(b)), spanning the full spectrum of  anomalies:     \textbf{(1) Perception Failure (PF):} Arises from the challenging optical properties of materials. It includes \textit{Target Position Deviation} and failures caused by transparent glassware or reflective fluid surfaces.
    \textbf{(2) Grasping Failure (GF):} Reflects end-effector control deficiencies, such as \textit{Object Slippage} due to varying friction coefficients of wet surfaces or insufficient gripping force.
    \textbf{(3) Motion Failure (MF):} Corresponds to trajectory execution anomalies. A domain-specific highlight is \textit{Incomplete Trajectory}, where premature termination of motion sequences leads to experimental failure.
    \textbf{(4) Logic Failure (LF):} Captures semantic planning errors (e.g., \textit{Step Omission}, \textit{Sequence Reversal}) that invalidate experimental outcomes.
    \textbf{(5) Safety Failure (SF):} This category covers \textit{Improper Handling} and \textit{Protocol Violations} (e.g., hazardous mixtures). Unlike reversible household failures, these pose risks of catastrophic damage or injury in chemical settings.

% \textbf{Paired Contrastive Data.} As illustrated in Fig.~\ref{fig:data}(d), for each task instance, we generate \textbf{paired success and failure trajectories} under identical initial environmental conditions. This rigorous alignment minimizes background confounding factors, enabling models to learn the precise discriminative boundaries between normal operations and subtle anomalies through contrastive supervision.
\textbf{Paired Contrastive Data.} We generate paired success and failure trajectories under identical conditions (Fig.~\ref{fig:data}(d)). This rigorous alignment minimizes confounds, facilitating the learning of discriminative anomalous boundaries.

\subsection{LabRobFail-Bench}
\label{sec:bench}

To comprehensively evaluate robotic resilience, we introduce \textbf{LabRobFail-Bench}, the first multi-dimensional benchmark tailored for the high-stakes environment of chemical laboratories. As illustrated in Fig.~\ref{fig:data}(c), the benchmark assesses agents across \textbf{six evaluation dimensions} organized into \textbf{three progressive cognitive levels}: (1) \textit{L1: Task Understanding}, (2) \textit{L2: Failure Detection \& Localization}, and (3) \textit{L3: Failure Analysis \& Correction}.

\textbf{Problem Formulation.} 
We define the failure analysis task as a mapping from multi-modal observations to a structured diagnostic report. Formally, given an input  $\mathcal{X} = \langle \mathcal{V}, \mathcal{Q} \rangle$ (where $\mathcal{V}$ denotes the RGB video stream, $\mathcal{Q}$ is the task instruction), the model predicts a composite failure state $\mathcal{Y}$:
\begin{equation}
    \mathcal{Y} = \langle \underbrace{y_{task}}_{\text{L1}}, \underbrace{y_{dete}, y_{loca}}_{\text{L2}}, \underbrace{y_{type}, y_{risk}, y_{corr}}_{\text{L3}} \rangle
\end{equation}
encompassing task status $y_{task}$, detection flags $y_{dete}$, temporal localization $y_{loca}$, failure type $y_{type}$, risk severity $y_{risk}$, and actionable correction policies $y_{corr}$.

% \textbf{L1: Task Understanding (Q1).} 
% This foundational level requires models to temporally decompose continuous video streams into atomic action sequences. The goal is to identify primitive operations (e.g., \textit{pick, place, pour, shake}) and their temporal dependencies. This provides a structured task representation, serving as the necessary context for subsequent anomaly reasoning.
\textbf{L1: Task Understanding (Q1).} Decomposes videos into atomic action sequences (e.g., \textit{pick, pour}) and their temporal dependencies, providing structured context for subsequent anomaly reasoning.

\textbf{L2: Failure Detection \& Localization (Q2-Q3).} Evaluates anomaly perception. \textbf{Q2} performs binary detection, while \textbf{Q3} localizes the failure to a specific frame number. Precise localization captures subtle transitions, enabling timely intervention before irreversible hazards occur.

% \textbf{L3: Failure Analysis \& Correction (Q4-Q6).} 
% This level assesses deep reasoning and recovery planning.
% \textbf{Q4 (Severity Assessment):} Categorizes failures into four hierarchical risk levels: (i) \textit{Minor} (negligible deviations), (ii) \textit{Recoverable} (requires step re-execution), (iii) \textit{Critical} (requires manual intervention), and (iv) \textit{Catastrophic} (involving safety violations or equipment damage).
% \textbf{Q5 (Type Classification):} Identifies the failure mechanism among perception, grasping, motion, logic, or safety categories to inform the recovery strategy.
% \textbf{Q6 (Correction Generation):} The core contribution of this benchmark. Unlike coarse suggestions in prior works, Q6 requires models to output \textbf{fine-grained, executable instructions}. These must explicitly specify arm trajectory adjustments, end-effector orientation ($\Delta R$), and gripper states, ensuring direct applicability for closed-loop recovery.
\textbf{L3: Failure Analysis \& Correction (Q4-Q6).} Targets deeper reasoning. \textbf{Q4} assigns failures to four severity levels: \textit{Minor}, \textit{Recoverable}, \textit{Critical}, and \textit{Catastrophic}. \textbf{Q5} classifies the failure type, while \textbf{Q6} outputs fine-grained, executable corrections over trajectory, orientation ($\Delta R$), and gripper states to support closed-loop recovery.

\section{LabRobFail-VLM}
\label{sec:LabRobFail_vlm}

We develop \textbf{LabRobFail-VLM}, a specialized vision-language model for laboratory failure analysis, as shown in Fig.~\ref{fig:vlm}. Rather than reducing failure recognition to binary classification, LabRobFail-VLM formulates it as a structured, recovery-oriented prediction task that jointly supports task understanding, failure detection and localization, failure diagnosis, and corrective guidance.

\subsection{Spatio-Temporal Input Representation}

\begin{table*}[t]
\centering
\vspace{-3pt}
\def\arraystretch{0.8}
\caption{Main results on LabRobFail-Bench (Seen Environments). Q1-Q5 denote Task Understanding, Failure Detection, Temporal Localization, Severity Assessment, and Failure Classification (accuracy \%). Q6 denotes Correction Strategy (BLEU and ROUGE-L). Best results are in \textbf{bold}.}
\label{tab:main_seen}
\resizebox{\textwidth}{!}{
\begin{tabular}{l|ccccc|ccccc}
\toprule
\multirow{2}{*}{\textbf{Method}} & \multicolumn{5}{c|}{\textbf{Accuracy} (\%) $\uparrow$} & \multicolumn{5}{c}{Q6: \textbf{Correction Strategy} $\uparrow$} \\
& Q1 & Q2 & Q3 & Q4 & Q5 & BLEU-1 & BLEU-2 & BLEU-3 & BLEU-4 & ROUGE-L \\
\midrule
Qwen3-VL-8B~\cite{qwen3vl} & 47.45 & 59.50 & 5.59 & 47.84 & 35.88 & 0.2594 & 0.1249 & 0.0626 & 0.0366 & 0.2976 \\
LLaVA-1.6-13B~\cite{liu2024llavanext} & 35.80 & 52.96 & 13.24 & 33.53 & 31.18 & 0.2278 & 0.0999 & 0.0506 & 0.0255 & 0.2598 \\
InternVL2.5-8B~\cite{chen2024expanding} & 28.10 & 47.04 & 12.35 & 18.24 & 42.06 & 0.1174 & 0.0507 & 0.0258 & 0.0149 & 0.1430 \\
DeepSeek-VL2-small~\cite{wu2024deepseek} & 34.25 & 47.51 & 13.82 & 48.82 & 33.18 & 0.1451 & 0.0852 & 0.0311 & 0.0182 & 0.1818 \\
\midrule
Gemini-2.0-flash~\cite{comanici2025gemini} & 26.98 & 56.85 & 6.76 & 44.41 & 36.89 & 0.1161 & 0.0540 & 0.0282 & 0.0160 & 0.1440 \\
Gemini-2.5-flash~\cite{comanici2025gemini} & 38.92 & 57.13 & 15.53 & 48.84 & 39.68 & 0.1275 & 0.0852 & 0.0538 & 0.0245 & 0.1231 \\
GPT-5.4~\cite{openai2026gpt54} & 45.16 & 52.34 & 15.88 & 47.06 & 38.54 & 0.1194 & 0.0606 & 0.0307 & 0.0161 & 0.1520 \\
\midrule
LabRobFail-VLM (Ours) & \textbf{94.45} & \textbf{90.83} & \textbf{77.21} & \textbf{83.18} & \textbf{73.21} & \textbf{0.7629} & \textbf{0.7569} & \textbf{0.7452} & \textbf{0.7362} & \textbf{0.7542} \\
\bottomrule
\end{tabular}
}
\vspace{-8pt}
\end{table*}

Standard VLMs struggle with long-horizon, multi-view robotic videos due to excessive visual tokens. We therefore represent each trajectory as a compact \textit{Spatio-Temporal Keyframe Grid} that preserves both temporal progression and synchronized multi-view observations.
Formally, let $\mathcal{V} = \{V^{(k)}\}_{k=1}^{K}$ denote synchronized video streams from $K$ viewpoints, where $V^{(k)} = \{I^{(k)}_1, \dots, I^{(k)}_T\}$ is the frame sequence captured from the $k$-th view. We first sample $M$ key timestamps $\{t_1, \dots, t_M\}$ according to action transitions, so that the selected frames summarize the critical stages of the manipulation process. We then construct an image grid $X_{\text{grid}} \in \mathbb{R}^{H \times W \times 3}$ by arranging the sampled frames into an $M \times K$ matrix, where rows correspond to temporal steps and columns correspond to viewpoints. We also render the temporal index $t$ at the top-left of each image, allowing explicit time reference during failure localization across views.

\subsection{Architecture and Hybrid Fine-tuning}

In Fig.~\ref{fig:vlm}, LabRobFail-VLM is built upon Qwen3-VL~\cite{qwen3vl} and adapted to the laboratory failure-analysis setting through a hybrid fine-tuning. The model consists of a vision encoder $E_{\phi}$, a modality projector $P_{\psi}$, and a  language  decoder $D_{\theta}$. Given an input tuple $\mathcal{X} = (X_{\text{grid}}, \mathcal{Q}_{\text{instruct}})$, it first extracts visual and textual representations as:
\begin{equation}
Z_v = P_{\psi}(E_{\phi}(X_{\text{grid}})), \qquad
Z_t = \text{Tokenizer}(\mathcal{Q}_{\text{instruct}}),
\end{equation}
and then autoregressively predicts the  failure output:
\begin{equation}
P(\mathcal{Y}\mid \mathcal{X}) = \prod_{j=1}^{L} D_{\theta}(y_j \mid y_{<j}, Z_v, Z_t),
\end{equation}
where $\mathcal{Y}$ is the state of composite failure defined in Eq.~(3).

\paragraph{Hybrid Fine-tuning Strategy.}
% A key challenge is the domain gap between web-scale pretraining and laboratory observations, where transparent glassware, reflective liquids, and subtle manipulation errors are underrepresented. To improve domain adaptation while preserving language reasoning, we adopt an asymmetric optimization strategy.
A key challenge is the domain gap between web-scale pretraining and laboratory scenes, where transparent glassware, reflective liquids, and subtle errors are underrepresented. We address this with asymmetric optimization that improves domain adaptation while preserving language reasoning.

Specifically, we divide the model parameters into vision-related and language-related components $\Theta_V = \{\phi, \psi\}$  and $\Theta_L = \{\theta\}$. We apply \textit{full fine-tuning} to $\Theta_V$ so that the visual encoder and projector can better capture laboratory-specific perceptual cues. In contrast, for $\Theta_L$, we employ LoRA~\cite{hu2022lora} to adapt the language model to our structured diagnostic and correction format while preserving its general reasoning capability. The resulting training objective is
\begin{equation}
\label{eq:objective}
\min_{\phi, \psi, \Delta\theta}
\sum_{(\mathcal{X}, \mathcal{Y}) \in \mathcal{D}}
-\log P(\mathcal{Y} \mid \mathcal{X}; \phi, \psi, \theta_0 + \Delta\theta),
\end{equation}
where $\theta_0$ denotes the frozen pretrained model parameters, and $\Delta\theta$ is the low-rank adaptation update. This hybrid strategy enables LabRobFail-VLM to strengthen domain-specific visual perception while retaining stable high-level reasoning for structured failure diagnosis and correction.

\section{Experiment}

\begin{table*}[t]
\centering
\vspace{-3pt}
\def\arraystretch{0.8}
\caption{Generalization to unseen environments: novel objects (Object), novel scene backgrounds (Scene), and both (Both).}
\label{tab:main_unseen}
 % \fontsize{8.0}{15.5}\selectfont
\resizebox{\textwidth}{!}{
\begin{tabular}{l|l|ccccc|ccccc}
\toprule
\multirow{2}{*}{\textbf{Method}} & \multirow{2}{*}{\textbf{Setting}} & \multicolumn{5}{c|}{\textbf{Accuracy} (\%) $\uparrow$} & \multicolumn{5}{c}{Q6:\textbf{ Correction Strategy} $\uparrow$} \\
& & Q1 & Q2 & Q3 & Q4 & Q5 & BLEU-1 & BLEU-2 & BLEU-3 & BLEU-4 & ROUGE-L \\
\midrule
\multirow{3}{*}{Qwen3-VL-8B~\cite{qwen3vl}} 
& Object & 58.24 & 64.86 & 5.29 & 47.18 & 26.20 & 0.2183 & 0.0885 & 0.0381 & 0.0246 & 0.2731 \\
& Scene & 61.75 & 68.27 & 4.44 & 39.50 & 35.65 & 0.2474 & 0.1279 & 0.0707 & 0.0384 & 0.2748 \\
& Both & 56.09 & 62.83 & 4.75 & 43.28 & 30.66 & 0.2196 & 0.0717 & 0.0298 & 0.0201 & 0.2426 \\
\midrule
\multirow{3}{*}{LLaVA-1.6-13B~\cite{liu2024llavanext}} 
& Object & 38.28 & 53.12 & 15.11 & 34.18 & 33.75 & 0.2331 & 0.0974 & 0.0527 & 0.0266 & 0.2459 \\
& Scene & 52.54 & 61.45 & 15.81 & 38.37 & 29.13 & 0.2119 & 0.0901 & 0.0459 & 0.0234 & 0.2317 \\
& Both & 39.12 & 55.24 & 14.21 & 36.11 & 30.14 & 0.2108 & 0.0852 & 0.0512 & 0.0257 & 0.2427 \\
\midrule
\multirow{3}{*}{Gemini-2.5-flash~\cite{comanici2025gemini}} 
& Object & 38.12 & 56.27 & 12.53 & 41.24 & 34.34 & 0.1214 & 0.0524 & 0.0425 & 0.0241 & 0.1524 \\
& Scene & 39.24 & 59.21 & 14.24 & 40.24 & 38.98 & 0.1304 & 0.0545 & 0.0398 & 0.0197 & 0.1672 \\
& Both & 39.51 & 56.54 & 11.12 & 38.54 & 36.54 & 0.1245 & 0.0721 & 0.0456 & 0.0232 & 0.1587 \\
\midrule
\multirow{3}{*}{LabRobFail-VLM (Ours)} 
& Object & \textbf{91.53} & \textbf{79.15} & \textbf{60.22} & \textbf{53.05} & \textbf{59.86} & \textbf{0.4876} & \textbf{0.4536} & \textbf{0.3974} & \textbf{0.3492} & \textbf{0.4668} \\
& Scene & \textbf{93.42} & \textbf{85.56} & \textbf{74.39} & \textbf{63.51} & \textbf{72.63} & \textbf{0.6793} & \textbf{0.6276} & \textbf{0.5805} & \textbf{ 0.5229} & \textbf{0.6604} \\
& Both & \textbf{89.09} & \textbf{71.02} & \textbf{41.41} & \textbf{46.02} & \textbf{48.72} & \textbf{0.3981} & \textbf{0.3875} & \textbf{0.3735} & \textbf{0.3269} & \textbf{0.3894} \\
\bottomrule
\end{tabular}
}
\vspace{-8pt}
\end{table*}

\begin{table}[t]
\centering
\small
\setlength{\tabcolsep}{4pt}
\caption{Ablation study on the Seen split. Accuracy (\%) for Q2, Q3 and Q5; ROUGE-L for Q6.}
\label{tab:ablation}
\begin{tabular}{lcccc}
\toprule
Variant & Q2 & Q3 & Q5 & Q6 \\
\midrule
Full fine-tuning   & 85.24 & 72.24 & 70.24 & 0.7142 \\
Frozen vision      & 81.13 & 60.21 & 61.24 & 0.4529 \\
w/o temporal index & 86.45 & 70.58 & 69.24 & 0.6471 \\
\midrule
LabRobFail-VLM & \textbf{90.83} & \textbf{77.21} & \textbf{73.21} & \textbf{0.7542} \\
\bottomrule
\end{tabular}
\vspace{-4pt}
\end{table}

\begin{table}[t]
\def\arraystretch{0.55}
\centering
\small
% \vspace{-3pt}
\caption{Success rate of downstream Policy recovery tasks.}
\label{tab:vla}
\scalebox{0.8}{
\begin{tabular}{l|lc|cc|c}
\toprule
\textbf{Task} & \textbf{Baseline} & \textbf{Rate(\%)} & \textbf{+Ours} & \textbf{Rate(\%)} & \textbf{Improv.} \\
\midrule
Pick & OpenVLA & 76 & 21/25 & 84 & \textbf{+8\%} \\
Place & OpenVLA & 48 & 13/25 & 52 & \textbf{+4\%} \\
Pour & OpenVLA & 16 & 7/25 & 28 & \textbf{+12\%} \\
Open drying oven & OpenVLA & 28 & 10/25 & 40 & \textbf{+12\%} \\
Close drying oven & OpenVLA & 12 & 5/25 & 20 & \textbf{+8\%} \\
\midrule
Pick & ACT & 64 & 17/25 & 68 & \textbf{+4\%} \\
Place & ACT & 52 & 15/25 & 60 & \textbf{+8\%} \\
Pour & ACT & 32 & 12/25 & 48 & \textbf{+16\%} \\
\bottomrule
\end{tabular}}
\vspace{-4pt}
\end{table}

\begin{table}[t]
% \vspace{-10pt}
\def\arraystretch{0.65}
\centering
\small
% \vspace{-5pt}
\caption{Effect of pretraining on LabRobFail-Data.}
\label{tab:transfer}
\begin{tabular}{lcc}
\toprule
% \multirow{2}{*}{\textbf{Dataset}} & \multicolumn{2}{c}{\textbf{Accuracy} (\%) $\uparrow$} \\
% \cmidrule(lr){2-3}
% \multirow{2}{*}{\textbf{Dataset}} & \multicolumn{2}{c}{\textbf{Accuracy} (\%) $\uparrow$} \\
% \cmidrule(lr){2-3}
Dataset & w/o Aux & w/ Aux \\
\midrule
AHA & 59.87 & 68.85 \\
AHA + LabRobFail-Data (Ours) & \textbf{63.96} & \textbf{72.74} \\
\bottomrule
\end{tabular}
\vspace{-8pt}
\end{table}

\subsection{Experimental Setup}

% \noindent \textbf{Evaluation Metrics.} 
% We employ tailored metrics for the six evaluation dimensions in LabRobFail-Bench. For Task Understanding (Q1), Failure Detection (Q2), Temporal Localization (Q3), Severity Assessment (Q4), and Failure Classification (Q5), we report \textbf{Top-1 Accuracy}. For Correction Strategy Generation (Q6), we adopt standard NLP metrics including \textbf{BLEU-n} ($n=1\dots4$) and \textbf{ROUGE-L} to quantify the lexical alignment and semantic consistency between the generated corrections and ground-truth references.
\noindent \textbf{Implementation Details.} 
To address the significant domain gap in laboratory scenarios, we adopt the hybrid fine-tuning strategy on Qwen3-VL-8B~\cite{qwen3vl}. The vision encoder and the projector are fully fine tuned with learning rates of $2\times10^{-6}$ and $1\times10^{-5}$, while the language decoder is adapted with LoRA~\cite{hu2022lora} of rank 64, scaling factor 128, dropout 0.05 and a learning rate of $1\times10^{-4}$. We train for 3 epochs with a global batch size of 256. The optimization uses a cosine scheduler (warmup 0.03) on 8 NVIDIA H100 GPUs via DeepSpeed ZeRO-3.

\noindent \textbf{Evaluation Metrics.} 
We report \textit{Top-1 Accuracy} for tasks Q1--Q5 (task understanding, detection, localization, severity, and classification). For correction (Q6), we employ \textit{BLEU-n} ($n=1\dots4$) and \textit{ROUGE-L} to quantify lexical alignment and semantic consistency with ground-truth references.

\subsection{Quantitative Experimental Results}

\begin{figure}[t]
\centering
\includegraphics[width=1.0\columnwidth]{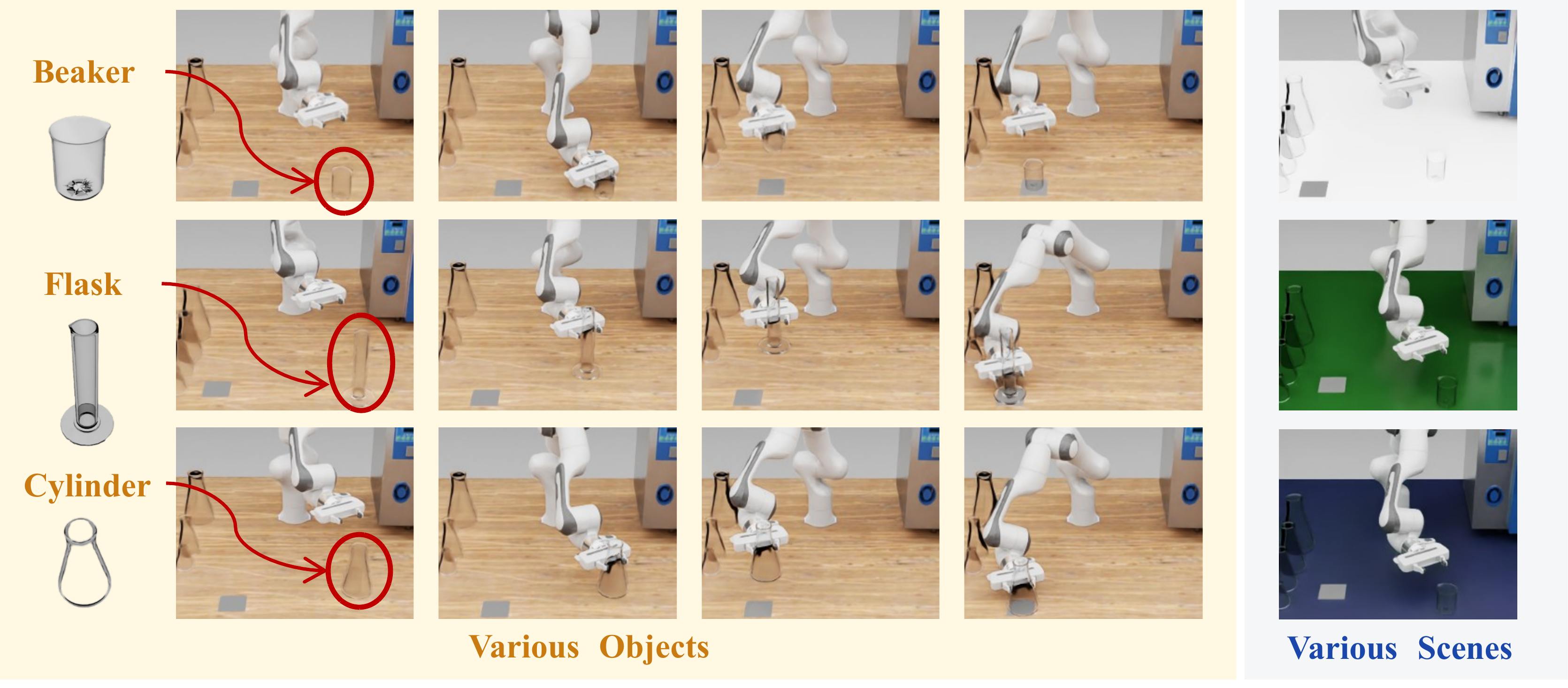}
\caption{Examples of unseen objects and scenes.}
\label{fig:seen}
\vspace{-10pt}
\end{figure}

To comprehensively evaluate model performance and robustness, we partition the test set into two distinct subsets: \textbf{Seen} (overlapping with training domains) and \textbf{Unseen} (containing novel scenes and target objects). Fig.~\ref{fig:seen} shows examples of objects and scenes used in the Unseen evaluation setting.

% \textbf{Performance on Seen Environments.}  LabRobFail-VLM demonstrates an overwhelming performance advantage across all six evaluation dimensions, validating the necessity of domain-specific adaptation.  Notably, on the critical tasks of failure detection (Q2) and temporal localization (Q3), our model achieves accuracies of \textbf{92.58\%} and \textbf{85.58\%}, respectively. This significantly outperforms generalist baselines, establishing a reliable foundation for anomaly identification. 
%  The model further ensures diagnostic reliability with over \textbf{90\%} accuracy in reasoning tasks (Q4-Q5), far exceeding the $\sim$68\% baseline ceiling.
% Furthermore, in correction strategy generation (Q6), LabRobFail-VLM attains high semantic alignment with expert demonstrations, reaching \textbf{0.74} in BLEU-4 and \textbf{0.89} in ROUGE-L. This confirms that without domain-specific fine-tuning, general-purpose models are incapable of synthesizing executable, scientific instructions.
% \noindent \textbf{Performance on Seen Environments.} LabRobFail-VLM consistently outperforms all baselines across the six evaluation dimensions. It achieves \textbf{90.83\%} accuracy in failure detection (Q2) and \textbf{77.21\%} in temporal localization (Q3), while exceeding \textbf{70\%} on severity assessment and failure classification (Q4–Q5). For correction generation (Q6), it reaches \textbf{0.74} BLEU-4 and \textbf{0.75} ROUGE-L, demonstrating the importance of domain-specific adaptation for reliable failure diagnosis and actionable correction.
\noindent \textbf{Performance on Seen Environments.} LabRobFail-VLM consistently outperforms all baselines across six evaluation dimensions, achieving \textbf{90.83\%} on failure detection (Q2), \textbf{77.21\%} on temporal localization (Q3), and over \textbf{70\%} on severity assessment and failure classification (Q4--Q5). It also obtains \textbf{0.74} BLEU-4 and \textbf{0.75} ROUGE-L for correction generation (Q6), highlighting the value of domain-specific adaptation for accurate diagnosis and actionable recovery.

\noindent \textbf{Generalization to Unseen Environments.} 
Table~\ref{tab:main_unseen} reports generalization under three progressively harder settings: novel objects (\textit{Object}), novel scenes (\textit{Scene}), and both combined (\textit{Both}). LabRobFail-VLM consistently outperforms generalist VLMs across all settings. In the challenging \textit{Both} setting, LabRobFail-VLM achieves \textbf{71.02\%} on Q2, surpassing Gemini-2.5-flash (56.54\%) and Qwen3-VL-8B (62.83\%), while retaining markedly better Q6 correction quality (BLEU-4: 0.3269 vs.\ at most 0.0257). However, combined shifts still reduce Q3 from 74.39\% to 41.41\% and Q5 from 72.63\% to 48.72\%, leaving robustness under compound distribution shifts an open challenge.

% \textbf{Effectiveness of Fine-tuning.} Compared to its base model Qwen3-VL-8B~\cite{qwen3vl}, LabRobFail-VLM shows substantial improvements, with Q2 (failure detection) accuracy increasing from 59.50\% to 92.58\% on Seen environments and from 62.83\% to 79.54\% on Unseen (Both) environments. This validates the necessity of domain-specific fine-tuning and demonstrates the effectiveness of LabRobFail-Data.

\subsection{Ablation Studies}
% Table~\ref{tab:ablation} ablates the main design choices on the Seen split. Full fine-tuning degrades all four dimensions, while freezing the vision encoder and the projector gives the weakest variant, losing 17.00 points on Q3, which together support the asymmetric optimization in Eq.~(\ref{eq:objective}). Removing the rendered temporal index reduces Q3 by 6.63 points, the largest drop among the accuracy dimensions for this variant, and supports explicit time references as the basis for grounding at the frame level.
Table~\ref{tab:ablation} evaluates the key design choices on the Seen split. Full fine-tuning degrades all four metrics, while freezing both the vision encoder and projector performs worst, including a 17.00-point drop on Q3, validating the asymmetric optimization in Eq.~(\ref{eq:objective}). Removing temporal indices reduces Q3 by 6.63 points—the largest accuracy drop for this variant—confirming their importance for frame-level grounding.

\subsection{Downstream Policy Recovery}

% To evaluate closed-loop recovery, we integrate LabRobFail-VLM as an \textbf{external supervisor} of two downstream policies, OpenVLA~\cite{kim2024openvla} and ACT~\cite{zhao2023act}, as shown in Fig.~\ref{fig:vlm}(b). The supervisor monitors execution and generates fine-grained corrections for arm trajectories, gripper states, and execution conditions, which a deterministic \textbf{Action Dictionary} maps to executable control primitives without additional fine-tuning. As shown in Table~\ref{tab:vla}, success rates improve across all eight settings, by 4 to 16 percentage points, with the largest gain on \textit{Pour} under ACT (32\% to 48\%). Absolute success rates nevertheless remain low on the harder tasks, indicating that reliable recovery in laboratory settings is still an open problem.
To evaluate closed-loop recovery, we deploy LabRobFail-VLM as an \textbf{external supervisor} for OpenVLA~\cite{kim2024openvla} and ACT~\cite{zhao2023act}. It generates fine-grained corrections for arm motion, gripper state, and execution conditions, which a deterministic \textbf{Action Dictionary} converts into control primitives without additional fine-tuning. Table~\ref{tab:vla} shows gains of 4--16 percentage points across all eight settings, with the largest improvement on \textit{Pour} under ACT (32\% to 48\%). However, low absolute success rates on harder tasks indicate that reliable laboratory recovery remains an open challenge.

\subsection{Transferability Value of LabRobFail-Data}
\label{subsec:transfer}
 To validate the transferability of LabRobFail-Data, we pre-train the AHA model~\cite{duan2024aha} using LabRobFail-Data. Table~\ref{tab:transfer} shows consistent improvements: accuracy rises by \textbf{+4.1\%} (59.87\% $\to$ 63.96\%) in isolation and by \textbf{+3.9\%} (68.85\% $\to$ 72.74\%) even when combined with large-scale auxiliary datasets. This shows that LabRobFail-Data provides unique, complementary failure semantics (e.g., liquid dynamics) that act as critical prior knowledge to enhance generalizable robustness.

\section{Conclusion and Limitations}

\textbf{Conclusion.} We introduce LabRobFail, a comprehensive framework comprising a physics-based simulation platform(LabRobFail-Sim), a large-scale dataset of 20K+ trajectories (LabRobFail-Data), and a multi-dimensional benchmark (LabRobFail-Bench). Our specialized model, LabRobFail-VLM, outperforms generalist baselines in failure reasoning and facilitates robust closed-loop recovery in downstream manipulation tasks, paving the way for reliable autonomous discovery.
% %We present LabRobFail, a comprehensive framework for robotic failure analysis in Self-Driving Laboratories. Leveraging existing simulation platforms, we develop LabRobFail-Sim for automated failure injection and construct LabRobFail-Data with over 20K trajectories across 5 major categories and 11 fine-grained failure types. We further establish LabRobFail-Bench, the first benchmark spanning six evaluation dimensions for failure analysis and correction in laboratory environments. Experiments reveal that current generalist VLMs exhibit significant limitations in laboratory failure analysis, while LabRobFail-VLM, fine-tuned on LabRobFail-Data, effectively bridges this gap and enables closed-loop recovery when integrated with VLA models.

\textbf{Limitations.} While LabRobFail supports real-time failure detection and fine-grained correction, training relies entirely on synthetic data, which leaves a visual domain gap for real-world deployment, and the deterministic Action Dictionary restricts recovery flexibility. Future work will extend the VLA experiments and adapt the policy to interpret open-ended language corrections directly.

\bibliography{aaai2027}

\clearpage
\appendix

\section{LabRobFail-Sim}
\label{appendix:sim}

\subsection{LLM Annotation Pipeline}

\subsubsection{VQA Generation Module}

To ensure the large language model accurately understands our requirements, we first assign it a specific role through a system prompt, establishing its expertise in laboratory robotics annotation:

\begin{quote}
\textit{``You are an expert annotator for robotic manipulation tasks in chemical laboratory environments. For each dimension, randomly select ONE question variant from the provided options, then generate the corresponding answer based on the visual observation and metadata.''}
\end{quote}

Subsequently, we provide a standardized image description template to help the model understand the spatio-temporal structure of the input keyframe grid:

\begin{quote}
\textit{``The image consists of a series of sequential frames. Each row presents a different camera viewpoint, and each column corresponds to a later time step, with the time-step label shown in the top-left corner. These frames depict the robotic arm's motion during Task: \{task\_name\}. For each of the following six dimensions, select ONE question from the provided variants and answer it.''}
\end{quote}

\begin{table*}[t]
\centering
\caption{Question variants for each evaluation dimension. GPT5.4 randomly selects one variant per dimension during annotation.}
\label{tab:question_variants}
\resizebox{\textwidth}{!}{
\begin{tabular}{cl}
\toprule
\textbf{Dimension} & \textbf{Question Variants} \\
\midrule
\multirow{4}{*}{\textbf{Q1: Task Understanding}} 
& - ``What atomic actions does the current task consist of? Please choose from: grasp, place, open, close, pour, shake, insert, stir, and list them in execution order.'' \\
& - ``Break down the robot's operation into primitive actions. Available primitives: grasp, place, open, close, pour, shake, insert, stir.'' \\
& - ``Identify the sequence of manipulation primitives performed in this task. Options: grasp, place, open, close, pour, shake, insert, stir.'' \\
& - ``Which atomic actions are executed in this task and in what order? Choose from: grasp, place, open, close, pour, shake, insert, stir.'' \\
\midrule
\multirow{5}{*}{\textbf{Q2: Failure Detection}} 
& - ``Is there any error in this task? (Yes/No)'' \\
& - ``Does the robot successfully complete the intended operation? (Yes/No)'' \\
& - ``Did anything go wrong during the execution? (Yes/No)'' \\
& - ``Is there any anomaly or failure observed in this trajectory? (Yes/No)'' \\
& - ``Was this task executed without issues? (Yes/No)'' \\
\midrule
\multirow{5}{*}{\textbf{Q3: Temporal Localization}} 
& - ``At which frame does the failure occur? (Refer to the frame number in the upper left corner)'' \\
& - ``Looking at the frame indices, when exactly does the error happen?'' \\
& - ``Identify the timestep where the failure first manifests. (Use the frame number displayed)'' \\
& - ``At what point in the sequence does the anomaly appear? (Specify the frame number)'' \\
& - ``Which frame marks the onset of the failure?'' \\
\midrule
\multirow{4}{*}{\textbf{Q4: Severity Assessment}} 
& - ``What is the severity level of this error?'' \\
& - ``How severe is this failure in terms of laboratory safety?'' \\
& - ``Assess the risk level of this failure.'' \\
& - ``Rate the severity of this anomaly.'' \\
\midrule
\multirow{4}{*}{\textbf{Q5: Failure Classification}} 
& - ``What is the failure type? Select the most appropriate category.'' \\
& - ``How would you classify the root cause of this failure?'' \\
& - ``Which category best describes this anomaly?'' \\
& - ``Identify the failure mode from the following categories.'' \\
\midrule
\multirow{5}{*}{\textbf{Q6: Correction Strategy}} 
& - ``What is the correction solution for this failure?'' \\
& - ``How should the robot recover from this situation?'' \\
& - ``Propose a specific corrective action to address this failure.'' \\
& - ``What adjustments should be made to fix this error?'' \\
& - ``Describe the recovery procedure for this anomaly.'' \\
\bottomrule
\end{tabular}
}
\end{table*}

Regarding question design, we provide multiple question variants for each evaluation dimension, as shown in Table~\ref{tab:question_variants}, to ensure diversity in the generated QA pairs. During annotation, GPT-5.4 randomly selects one variant per dimension, then generates the corresponding answer according to our provided configuration file.

For the answer options, we define fixed categories for Q4 and Q5 to ensure annotation consistency, as shown in Table~\ref{tab:answer_options}. The model reads these predefined options from our configuration file and selects the most appropriate category based on the observed failure context.

Based on the configuration file, GPT5.4 generates complete question-answer pairs for each trajectory. We then use a post-processing script to convert the raw outputs into our standardized annotation format. The generated annotations are validated in two steps. A rule-based filter removes outputs with malformed structure, frame indices outside the valid range, or answers inconsistent with the simulation metadata. A random sample of the remaining annotations is then manually inspected, and QA pairs that fail the inspection are regenerated. The following shows an example of the final formatted QA pair:

% \begin{table}[H] 
% \centering
% \tiny
% \caption{Answer options for Q4 (Severity Assessment) and Q5 (Failure Classification).}
% \label{tab:answer_options}
% \resizebox{\linewidth}{!}{
% \begin{tabular}{cll}
% \toprule
% \textbf{Dim.} & \textbf{Option} & \textbf{Definition} \\
% \midrule
% \multirow{4}{*}{Q4} 
% & (1) Risk Level & Collision, liquid spillage, or potential falling \\
% & (2) Dangerous Level & Requires manual intervention to proceed \\
% & (3) Fatal Level & Equipment damage, must be manually intervened \\
% & (4) Management Incident & Non-compliant operations (e.g., unclosed cabinet) \\
% \midrule
% \multirow{5}{*}{Q5} 
% & (1) Perception Failure & Failed to locate target objects or positions \\
% & (2) Grasping Failure & Object slippage or unstable grasp \\
% & (3) Motion Failure & Motion execution incomplete \\
% & (4) Logic Failure & Incorrect order or omitted steps \\
% & (5) Safety Failure & Violating laboratory safety protocols \\
% \bottomrule
% \end{tabular}
% }
% \end{table}

\begin{table*}[t]
\centering
\caption{Answer options for Q4 (Severity Assessment) and Q5 (Failure Classification).}
\label{tab:answer_options}
\begin{tabular}{c p{5cm} p{7cm}}
\toprule
\textbf{Dim.} & \textbf{Option} & \textbf{Definition} \\
\midrule
\multirow{4}{*}{Q4} 
& (1) Risk Level & Collision, liquid spillage, or potential falling \\
& (2) Dangerous Level & Requires manual intervention to proceed \\
& (3) Fatal Level & Equipment damage, requiring manual intervention \\
& (4) Management Incident & Non-compliant operations (e.g., unclosed cabinet) \\
\midrule
\multirow{5}{*}{Q5} 
& (1) Perception Failure & Failed to locate target objects or positions \\
& (2) Grasping Failure & Object slippage or unstable grasp \\
& (3) Motion Failure & Motion execution incomplete \\
& (4) Logic Failure & Incorrect order or omitted steps \\
& (5) Safety Failure & Violating laboratory safety protocols \\
\bottomrule
\end{tabular}
\end{table*}

\begin{table*}[t]
\def\arraystretch{0.92}
\centering
\small
\caption{Fine-grained correction strategies organized by failure category. Placeholders in \{brackets\} are filled based on specific task context.}
\label{tab:correction_strategies}
\begin{tabular}{l|p{12.5cm}}
\toprule
\textbf{Category} & \textbf{Correction Instruction} \\
\midrule
\multirow{2}{*}{\textbf{Safety Failure}} 
& Keep the robotic arm position unchanged, maintain gripper state, and keep the arm horizontal to complete the experiment safely. \\
& After completing the operation, keep the gripper horizontal, move the robotic arm, and place the \{target\} steadily. \\
\midrule
\multirow{2}{*}{\textbf{Logic Failure}} 
& First move the robotic arm to grasp the \{target\} with gripper closed, then move to the \{destination\} and open the gripper to place. \\
& Pause current execution, return to perform the omitted \{action\}, then resume the remaining workflow. \\
\midrule
\multirow{4}{*}{\textbf{Motion Failure}} 
& Adjust the gripper orientation to horizontal, keep the gripper closed, then move the robotic arm forward to insert into the \{target\}. \\
& Keep the gripper closed and orientation unchanged, move the robotic arm side to side repeatedly until the liquid is fully stirred. \\
& Keep the gripper closed on the door handle and move the robotic arm \{direction\} until the door is fully \{open/closed\}. \\
& Move the robotic arm to align the \{source\} above the \{destination\}, keep the gripper closed, then progressively tilt the gripper until the liquid is fully poured. \\
\midrule
\multirow{3}{*}{\textbf{Grasping Failure}} 
& Keep the gripper open and move the robotic arm until the \{target\} is centered within the gripper, then close the gripper to secure the grasp. \\
& Move the robotic arm toward the \{target\}, adjust the gripper orientation to horizontal, keep the gripper open until centered, then close the gripper to grasp. \\
& Keep the gripper closed to secure the \{target\} and move the robotic arm to the \{destination\} while maintaining orientation horizontal. \\
\midrule
\multirow{4}{*}{\textbf{Perception Failure}} 
& Move the robotic arm to center the gripper on the door handle, close the gripper to grasp, then move \{direction\} to \{open/close\} the door. \\
& Move the robotic arm to position the \{target\} above the \{destination\}, keep the gripper orientation horizontal, then open the gripper to release. \\
& Move the robotic arm to align the gripper with the \{target\}, keep the gripper closed, then move forward until pressed. \\
& Open the gripper to release, reposition the robotic arm to center the \{target\}, then close the gripper firmly to secure. \\
\midrule
\textbf{Manual Operation} 
& The object has fallen or critical failure occurred, please correct it manually right away. \\
\bottomrule
\end{tabular}
\end{table*}

\begin{lstlisting}[language=json, basicstyle=\ttfamily\scriptsize, frame=single, breaklines=true]
{
  "id": "10_Q4",
  "image": "close_drying_oven_episode_0_erro.png",
  "conversations": [
    {
      "from": "human",
      "value": "<image>\nThe image consists of a series of sequential frames...What is the severity level of this error?\nOptions:\n(1) Risk Level\n(2) Dangerous Level\n(3) Fatal Level\n(4) Management Incident"
    },
    {"from": "gpt", "value": "2"}
  ]
}
\end{lstlisting}

\begin{lstlisting}[language=json, basicstyle=\ttfamily\scriptsize, frame=single, breaklines=true] 
{
  "id": "10_Q6",
  "image": "close_drying_oven_episode_0_erro.png",
  "conversations": [
    {
      "from": "human",
      "value": "<image>\nThe image consists of a series of sequential frames...What is the correction solution for this failure?"
    },
    {"from": "gpt", "value": "Grasp the door handle with the gripper, move the robotic arm forward to close the door."}
  ]
}
\end{lstlisting}

The remaining dimensions (Q1-Q3, Q5) follow the same format, with each question-answer pair stored as an independent training sample.

\subsubsection{Correction Strategy Formulation}

We design a set of fine-grained correction instructions that can be mapped to executable robot actions. Each instruction follows a unified template structure containing four components: position adjustment, orientation adjustment, gripper state, and execution condition. Table~\ref{tab:correction_strategies} presents the complete set of correction strategies organized by failure category.

\section{LabRobFail-Data}
\label{appendix:data}

LabRobFail-Data encompasses a comprehensive failure taxonomy covering 5 major categories (Safety, Logic, Motion, Grasping, and Perception) and 11 fine-grained failure types. To provide a clear understanding of each failure type, we present a representative example for each failure category below, along with the corresponding visualization results. Each example shows the keyframe sequence where the failure occurs. The 11 failure types include: (1) \textit{Protocol Violation} and \textit{Improper Handling} under Safety; (2) \textit{Sequence Reversal} and \textit{Step Omission} under Logic; (3) \textit{Pose Control Error} and \textit{Incomplete Trajectory} under Motion; (4) \textit{Gripper Control Failure} and \textit{Object Slippage} under Grasping; and (5) \textit{Target Positioning Deviation}, \textit{Operation Position Error}, and \textit{Grasp Target Misalignment} under Perception.

\subsection{Safety Failure}

\textbf{Protocol Violation.} The stirring task consists of the following sequence: first, the glass rod is grasped from the test tube rack, then transported to a position above the beaker, inserted into the beaker, and subsequently oscillated gently in a lateral manner. A Protocol Violation occurs during the insertion phase, where an angular misalignment prevents the glass rod from being correctly inserted into the beaker, resulting in insufficient stirring that violates laboratory operation protocols.

\begin{figure}[h]
\centering
\includegraphics[width=\linewidth]{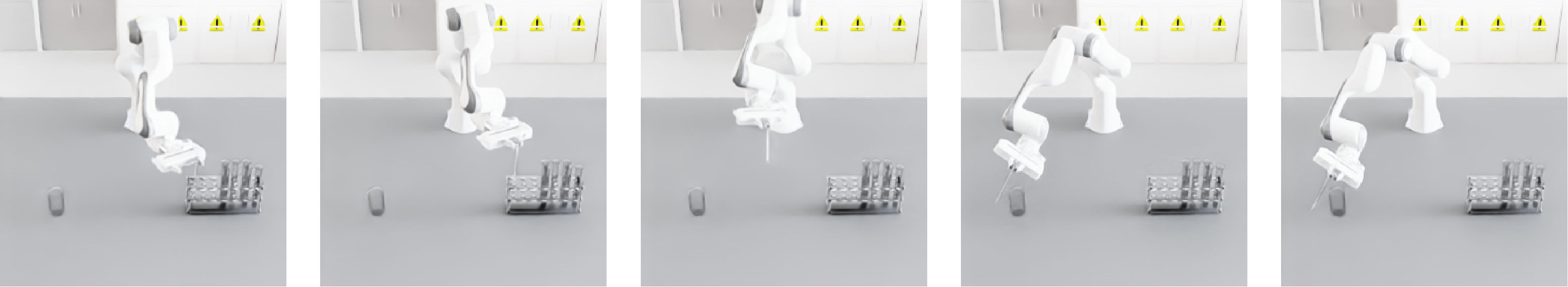}
\caption{Visualization of Protocol Violation failure.}
\label{fig:protocol_violation}
\end{figure}

\textbf{Improper Handling.} The pouring operation from the Erlenmeyer flask into the beaker consists of four sequential phases: grasping the flask, transporting it to a position above the beaker, executing the pouring motion, and subsequently returning the flask to its original location. An Improper Handling event occurs during the returning phase, wherein the flask fails to be placed back at its designated original position, violating the chemical laboratory requirement for proper equipment placement after use, thereby leading to an abnormal task outcome.

\begin{figure}[h]
\centering
\includegraphics[width=\linewidth]{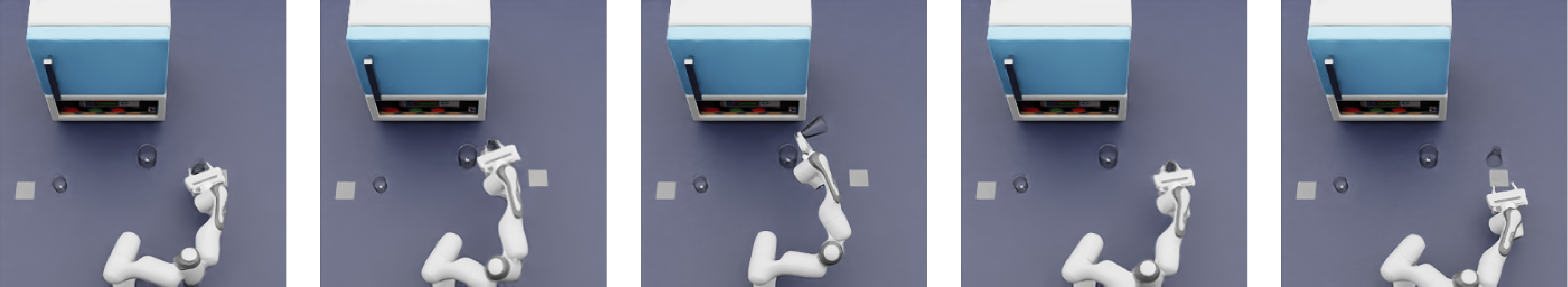}
\caption{Visualization of Improper Handling failure.}
\label{fig:improper_handling}
\end{figure}

\subsection{Logic Failure}

\textbf{Sequence Reversal.} The stirring task consists of the following sequence: first, the robotic arm grasps the glass rod from the test tube rack, then moves it to a position above the beaker, inserts the rod into the beaker, and performs a lateral stirring motion. A Sequence Reversal event occurs at the beginning of the task, where the robotic arm executes a stirring motion without the glass rod prior to grasping it, leading to an incorrect task execution.

\begin{figure}[h]
\centering
\includegraphics[width=\linewidth]{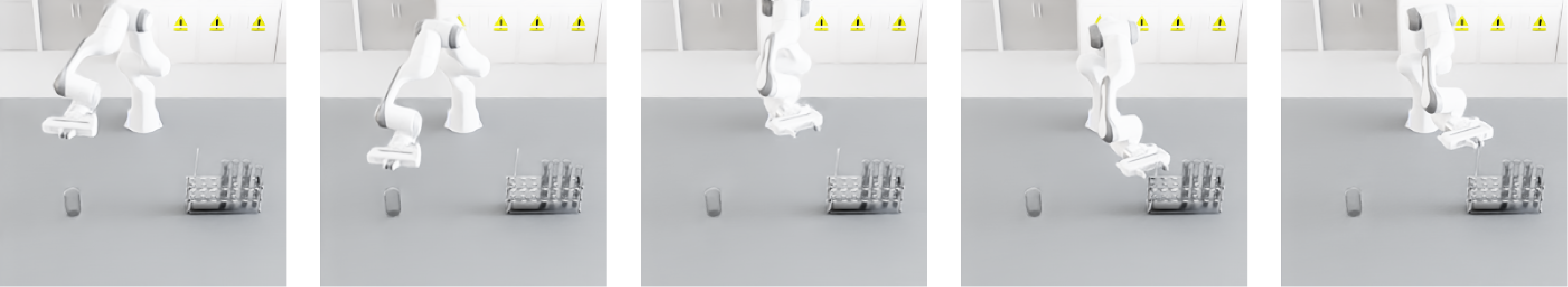}
\caption{Visualization of Sequence Reversal failure.}
\label{fig:sequence_reversal}
\end{figure}

\textbf{Step Omission.} The stirring task consists of the following sequence: first, the glass rod is grasped from the test tube rack, then transported to a position above the beaker, inserted into the beaker, and subsequently oscillated gently in a lateral manner. Step Omission occurs during the initial phase of the task when the robotic arm proceeded directly to stirring without gripping the glass rod.

\begin{figure}[H]
\centering
\includegraphics[width=\linewidth]{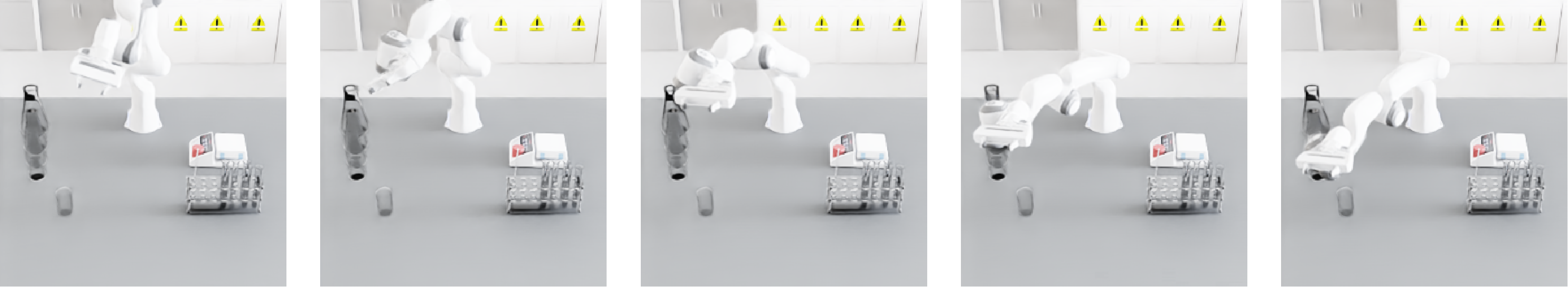}
\caption{Visualization of Step Omission failure.}
\label{fig:step_omission}
\end{figure}

\subsection{Motion Failure}

\textbf{Pose Control Error.} The placement task involves moving the beaker to a position above the target location. A Pose Control Error occurs during the beaker placement phase, where an abnormal twist in the robot arm's pose leads to unsuccessful placement of the beaker at the target location, resulting in task failure.

\begin{figure}[H]
\centering
\includegraphics[width=\linewidth]{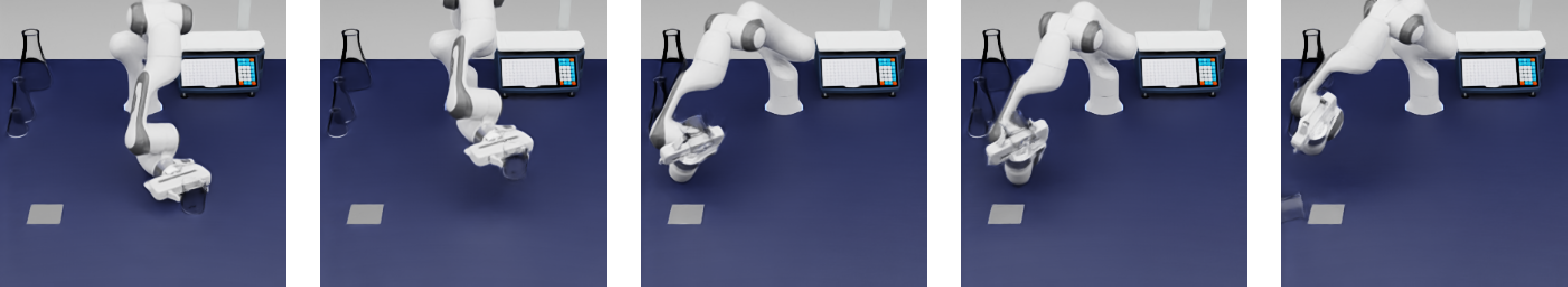}
\caption{Visualization of Pose Control Error failure.}
\label{fig:pose_control_error}
\end{figure}

\textbf{Incomplete Trajectory.} The button-pressing task requires the robotic arm to move to the designated red button and execute the press. An Incomplete Trajectory occurs during the approach phase: the arm stops prematurely and remains stationary in front of the button, causing task failure.

\begin{figure}[H]
\centering
\includegraphics[width=\linewidth]{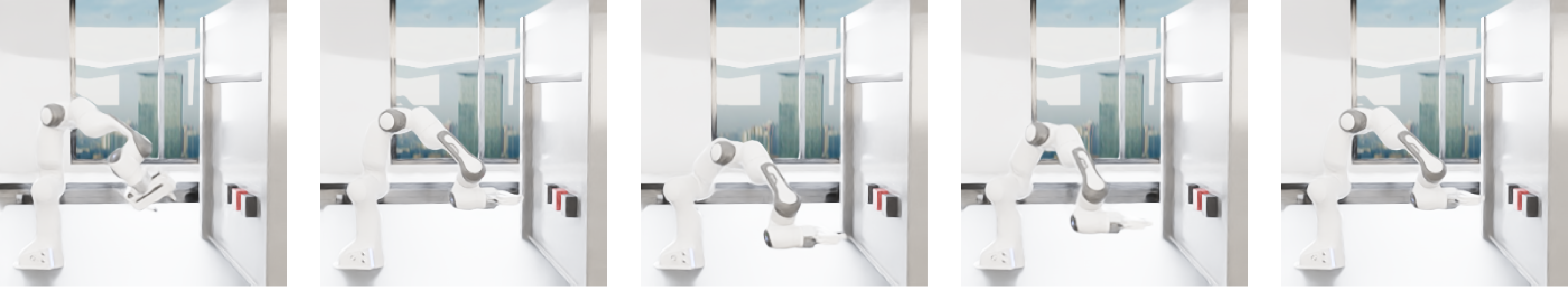}
\caption{Visualization of Incomplete Trajectory failure.}
\label{fig:incomplete_trajectory}
\end{figure}

\subsection{Grasping Failure}

\textbf{Gripper Control Failure.} The sequence begins with grasping the chemical instrument. A Gripper Control Failure manifests during the grasping phase. Inadequate gripper closure aperture prevents successful object acquisition, leading to task failure.

\begin{figure}[H]
\centering
\includegraphics[width=\linewidth]{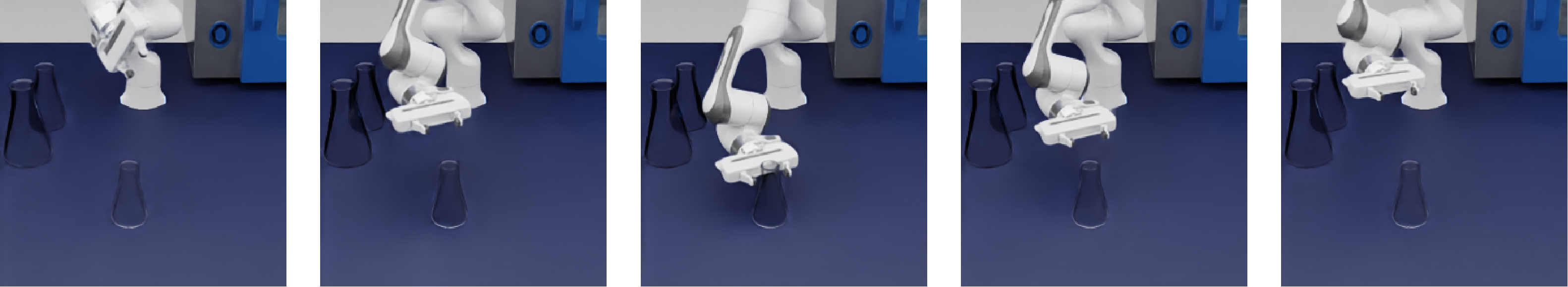}
\caption{Visualization of Gripper Control Failure.}
\label{fig:gripper_control_failure}
\end{figure}

\textbf{Object Slippage.} The placement task involves moving the chemical equipment to a designated location. The sequence consists of grasping the beaker, moving it above the target, and placing it at the target. An Object Slippage event occurs during the transport phase, where insufficient gripper force causes the beaker to slip and fall, resulting in task failure.

\begin{figure}[H]
\centering
\includegraphics[width=\linewidth]{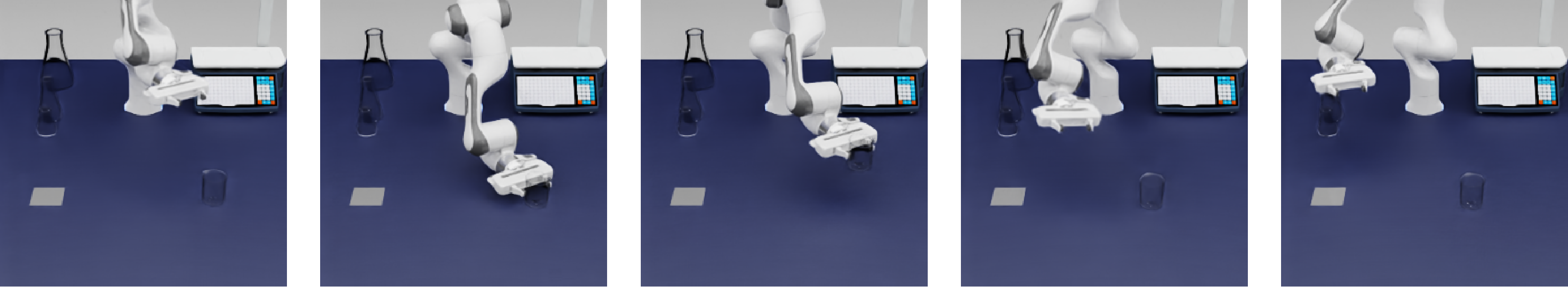}
\caption{Visualization of Object Slippage failure.}
\label{fig:object_slippage}
\end{figure}

\subsection{Perception Failure}

\textbf{Target Positioning Deviation.} The placement task involves the robotic arm first grasping the beaker and then placing it at the target location. A Target Positioning Deviation occurs during the final phase of the task, where the beaker is not placed within the designated target region, resulting in an abnormal outcome.

\begin{figure}[H]
\centering
\includegraphics[width=\linewidth]{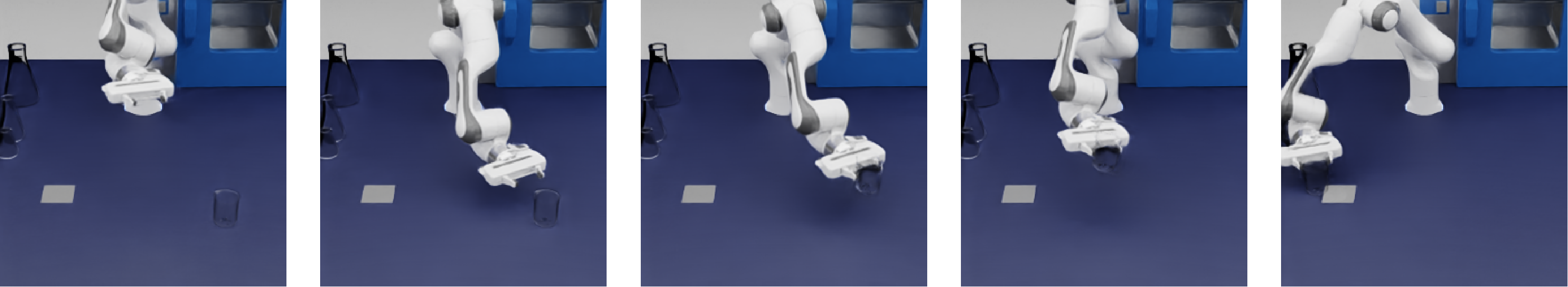}
\caption{Visualization of Target Positioning Deviation failure.}
\label{fig:target_positioning_deviation}
\end{figure}

\textbf{Operation Position Error.} The door-opening task involves moving the robotic arm to the front of the door handle, grasping the handle, and then pulling it backward in a fan-shaped motion. An Operation Position Error occurs when the gripper moves to the front of the door handle, where misalignment between the gripper and the handle position causes the door-opening operation to fail.

\begin{figure}[H]
\centering
\includegraphics[width=\linewidth]{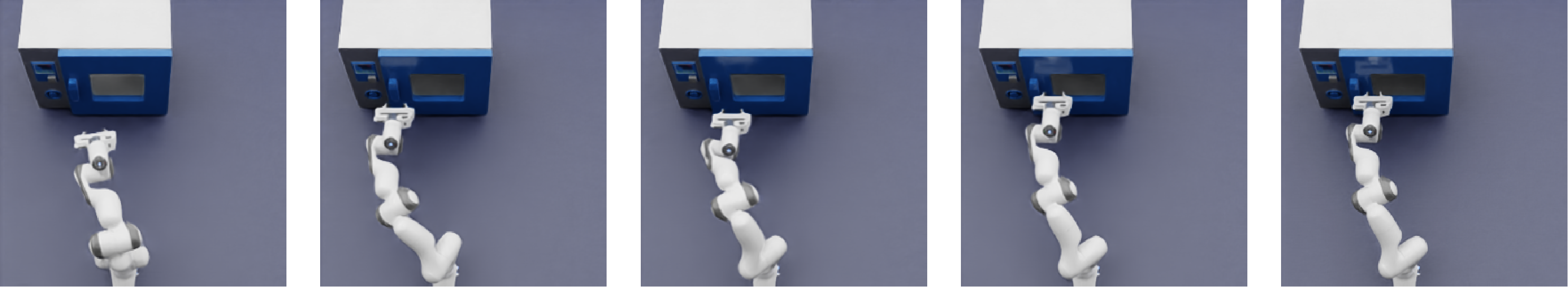}
\caption{Visualization of Operation Position Error failure.}
\label{fig:operation_position_error}
\end{figure}

\textbf{Grasp Target Misalignment.} The sequence begins with grasping the chemical instrument. A Grasp Target Misalignment manifests during the grasping phase: the estimated target position deviates from the actual object position, so the gripper closes beside the object and fails to acquire it, leading to task failure.

\begin{figure}[H]
\centering
\includegraphics[width=\linewidth]{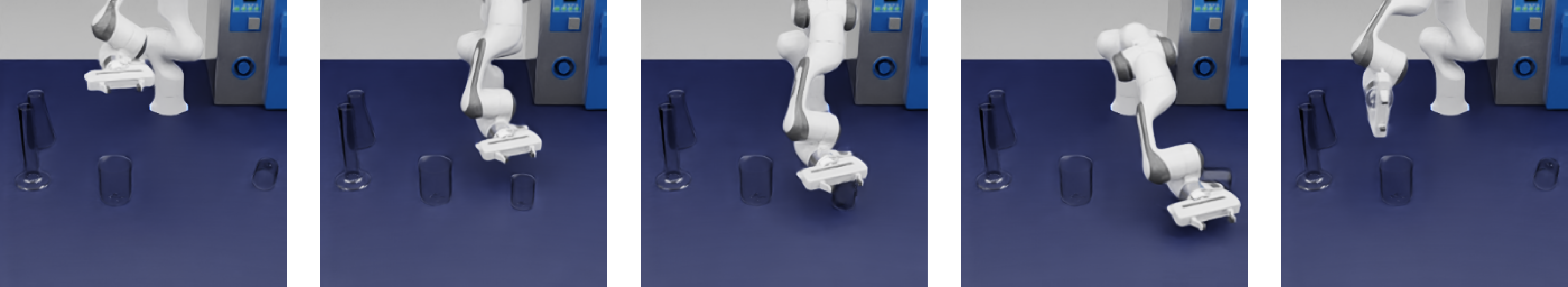}
\caption{Visualization of Grasp Target Misalignment failure.}
\label{fig:grasp_target_misalignment}
\end{figure}

\section{Experiment Details}
\label{appendix:downstream}

\subsection{Training Details}

We adopt Qwen3-VL-8B as the backbone and apply a hybrid fine-tuning strategy: full fine-tuning for the vision encoder and modality projector to adapt to laboratory-specific visual features (e.g., transparent glassware, reflective liquid surfaces), and LoRA for the LLM decoder to preserve general reasoning capabilities. The LoRA configuration uses rank $r=64$ and scaling factor $\alpha=128$. The input Spatio-Temporal Keyframe Grid has a resolution of 1536 $\times$ 768 pixels. We split LabRobFail-Data into training, validation, and test sets with a ratio of 8:1:1. For the Seen subset, we employ random splitting to divide the dataset. For the Unseen subset, we perform diversity expansion on target objects and scene backgrounds to ensure inconsistency with the training distribution. Other training configurations (learning rates, batch size, optimizer) follow the Implementation Details in the main paper.

\subsection{Action Dictionary}

\begin{table*}[h]
\def\arraystretch{0.92}
\centering
\small
\caption{Action Dictionary: Position and Orientation mappings.}
\label{tab:action_dict_position}
\begin{tabular}{p{5.5cm}|p{5.5cm}}
\toprule
\textbf{Semantic Instruction} & \textbf{Executable Action} \\
\midrule
\multicolumn{2}{c}{\textit{Position Actions}} \\
\midrule
keep the robotic arm position unchanged & \texttt{hold\_position()} \\
move the robotic arm forward & \texttt{delta\_pos(x=+d, y=0, z=0)} \\
move the robotic arm backward & \texttt{delta\_pos(x=-d, y=0, z=0)} \\
move the robotic arm \{direction\} & \texttt{delta\_pos(direction, dist)} \\
move the robotic arm side to side repeatedly & \texttt{oscillate(axis='y', amp, n)} \\
move the robotic arm to center the gripper on \{target\} & \texttt{align\_to(target\_id)} \\
move the robotic arm to position \{target\} above \{destination\} & \texttt{move\_above(dest\_id, h)} \\
move the robotic arm to align \{source\} above \{destination\} & \texttt{align\_above(src, dest)} \\
return to perform the omitted \{action\} & \texttt{move\_to\_pose(pose\_id)} \\
reposition the robotic arm to center the \{target\} & \texttt{align\_to(target\_id)} \\
\midrule
\multicolumn{2}{c}{\textit{Orientation Actions}} \\
\midrule
keep the arm horizontal & \texttt{set\_orient(roll=0, pitch=0)} \\
maintain orientation unchanged & \texttt{hold\_orientation()} \\
adjust the gripper orientation to horizontal & \texttt{set\_orient(roll=0, pitch=0)} \\
tilt the gripper & \texttt{rotate(axis='y', angle=$\theta$)} \\
progressively tilt until liquid is fully poured & \texttt{tilt\_pour(angle=$\theta$, v)} \\
\bottomrule
\end{tabular}
\end{table*}

\begin{table*}[h]
\def\arraystretch{0.92}
\centering
\small
\caption{Action Dictionary: Gripper action mappings.}
\label{tab:action_dict_gripper}
\begin{tabular}{p{5.5cm}|p{5.5cm}}
\toprule
\textbf{Semantic Instruction} & \textbf{Executable Action} \\
\midrule
maintain gripper state & \texttt{hold\_gripper()} \\
open the gripper & \texttt{gripper\_open()} \\
close the gripper & \texttt{gripper\_close()} \\
close the gripper to secure the grasp & \texttt{gripper\_close()} \\
close the gripper firmly to secure & \texttt{gripper\_close()} \\
open the gripper to release & \texttt{gripper\_open()} \\
\bottomrule
\end{tabular}
\end{table*}

To bridge the gap between LabRobFail-VLM's semantic correction instructions and executable robot control, we construct an Action Dictionary that maps natural language descriptions to low-level control primitives for the Franka robot. Table~\ref{tab:action_dict_position} and Table~\ref{tab:action_dict_gripper} present the complete mapping organized by action type.

It is worth noting that while some of our semantic outputs cannot be directly mapped to low-level robot actions (e.g., high-level descriptions involving target objects or conditional execution), they provide rich contextual and positional information that enables downstream Vision-Language-Action (VLA) models to interpret and execute the intended corrections. This design choice lays the foundation for future work on end-to-end policy adaptation, where VLA models can directly consume natural language correction instructions without requiring a deterministic action dictionary.


\end{document}